\pdfoutput=1
\documentclass[runningheads]{llncs}
\usepackage{graphicx}

\usepackage{mathtools}
\usepackage{subfigure}
\usepackage{booktabs}
\usepackage{pifont}
\usepackage{epstopdf}
\usepackage{tikz}
\usepackage{comment}
\usepackage{amsmath,amssymb} 
\usepackage{color}

\usepackage[accsupp]{axessibility}  
\usepackage{hyperref}

\begin{document}
\pagestyle{headings}
\mainmatter
\def\ECCVSubNumber{5297}  

\title{Graph Neural Network for Cell Tracking in Microscopy Videos} 

\titlerunning{Graph Neural Network for Cell Tracking in Microscopy Videos}
%
\author{Tal Ben-Haim\index{Ben-Haim, Tal} \and
Tammy Riklin Raviv\index{Riklin Raviv, Tammy}}
\authorrunning{T. Ben-Haim and T. Riklin Raviv}
%
\institute{School of Electrical and Computer Engineering, Ben-Gurion University\\
\email{\{benhait@post, rrtammy@ee\}.bgu.ac.il}}

\maketitle

\begin{abstract}
We present a novel graph neural network (GNN) approach for cell tracking in high-throughput microscopy videos. By modeling the entire time-lapse sequence as a direct graph where cell instances are represented by its nodes and their associations by its edges, we extract the entire set of cell trajectories by looking for the maximal paths in the graph. This is accomplished by several key contributions incorporated into an end-to-end deep learning framework. We exploit a deep metric learning algorithm to extract cell feature vectors that distinguish between instances of different biological cells and assemble same cell instances. We introduce a new GNN block type which enables a mutual update of node and edge feature vectors, thus facilitating the underlying message passing process. The message passing concept, whose extent is determined by the number of GNN blocks, is of fundamental importance as it enables the `flow' of information between nodes and edges much behind their neighbors in consecutive frames. Finally, we solve an edge classification problem and use the identified active edges to construct the cells' tracks and lineage trees.

We demonstrate the strengths of the proposed cell tracking approach by applying it to 2D and 3D datasets of different cell types, imaging setups, and experimental conditions. We show that our framework outperforms current state-of-the-art methods on most of the evaluated datasets. The code is available at our repository\footnote{For our repository, refer to \url{https://github.com/talbenha/cell-tracker-gnn}.}.
\end{abstract}

\section{Introduction}
\label{sec:intro}
Time-lapse microscopy imaging is a common tool for the study of biological systems and processes. However, probing complex and dynamic cellular events requires quantitative analysis at the single cell level of a huge amount of data, which far surpasses human annotators' abilities.
Therefore, automatic cell tracking which aims to identify and associate instances of the same biological cells and their offspring along microscopy sequences is an active field of study. 

The automatic construction of cell trajectories is a challenging problem since same-sequence cells often have similar visual traits yet individual cells may change their appearance or even divide over the course of time. In addition, a cell may temporarily occlude another, exit and reenter the frame field of view. On top of these, frequent mitotic events, the high cell density, high cell migration rate and low frame rate render cell tracking even more difficult.

Classical cell tracking methods can be split into cell association algorithms which solve the tracking problem in a temporarily local manner, and global frameworks which aim to solve cell tracking for an entire time-lapse microscopy sequence by simultaneous extraction of entire tracks. 
Frame-to-frame association algorithms connect cell instances in consecutive frames based on their visual traits and pose similarity~\cite{bensch2015cell,mavska2013segmentation} as well as cell motion prediction~\cite{amat2014fast,arbelle2018probabilistic}. Methods that look for globally optimal cell tracking solutions usually represent a microscopy sequence by a graph, encoding detected cell instances and their potential links to neighboring frames by the graph nodes and edges, respectively.
In most cases, linear programming and combinatorial algorithms were used to extract, multiple, non-overlapping cell trajectories~\cite{bise2011reliable,jug2016moral,kausler2012discrete,magnusson2015global,padfield2011coupled,rempfler2017efficient,rempfler2018tracing,schiegg2015graphical,schiegg2013conservation}.
\begin{figure*}[t]
\centerline{\includegraphics[scale=0.19,trim={0cm 0.0cm 0 0cm},clip]{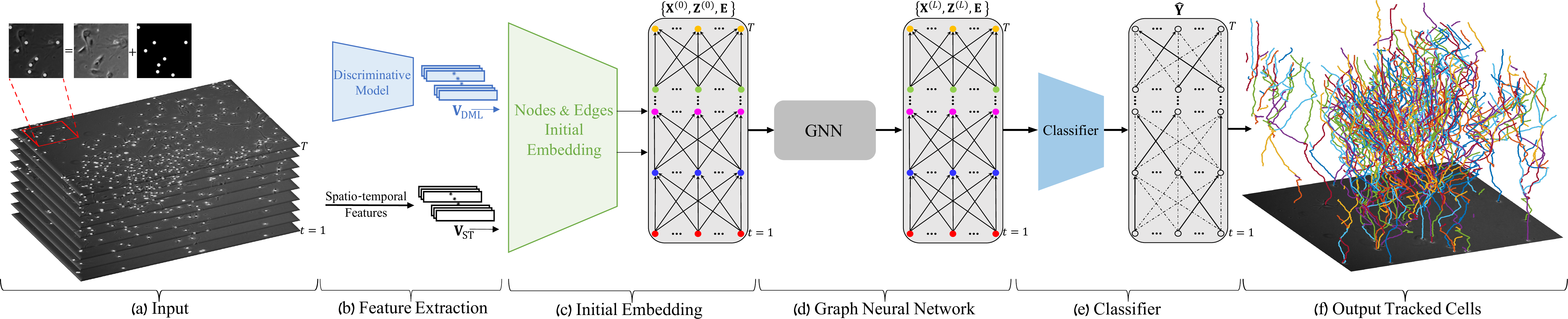}}
  \caption{\textbf{An outline of the proposed cell tracking framework.} (a) The input is composed of a live cell microscopy sequence of length $T$ and the corresponding sequence of label maps.
(b) Each cell instance in the sequence is represented by a feature vector which includes DML and spatio-temporal features. (c) The entire microscopy sequence is encoded as a direct graph where the cell instances are represented by its nodes and their associations are represented by the graph edges. Each node and edge in the graph has its own embedded feature vector. (d) These feature vectors are encoded and updated using Graph Neural Network (GNN). The GNN (which is illustrated in Fig.~\ref{fig:GNN}) is composed of $L$ message passing blocks which enable an update of edge and node features by their $L$-th order neighbors (i.e., cell instances which are up to $L$ frames apart). (e) The GNN's edge feature output is the input for an edge classifier network which classifies the edges into active (solid lines) and non-active (dashed lines). During training, the predicted classification $\hat{\textbf{Y}}$ is compared to the GT classification $\textbf{Y}$ for the loss computation. Since all the framework components are connected in an end-to-end manner the loss backpropogates throughout the entire network. (f) At inference time, cell tracks are constructed by concatenating sequences of active edges that connect cells in consecutive frames
\label{fig:model_diagram}} 
\end{figure*}

Despite their increasing popularity and state-of-the-art performance in various fields, deep learning approaches for tracing cells in a sequence became prevalent only lately. 
Many recent methods use convolutional neural networks (CNNs) for cell segmentation, but the construction of cell trajectories is still performed using classical methods e.g.,~\cite{hernandez2018cell,scherr2020cell,sixta2020coupling}. A pioneering complete deep learning method for cell tracking was proposed in~\cite{he2017cell} however it is limited to single cell tracking. 
In~\cite{hayashida2019cell} two separate U-nets~\cite{ronneberger2015u} were proposed for cell likelihood detection and movement estimation. In a later work by that group, a single U-Net was exploited to perform both tasks at once~\cite{Hayashida_2020_CVPR}. A further elaboration was proposed in~\cite{nishimura2020weakly} using both backward and forward propagation between nearby frames.
Both~\cite{payer2019segmenting} and \cite{spilger2020recurrent} presented recurrent neural network (RNN) approaches for the time-series analysis of microscopy videos. Other recent cell tracking approaches include a deep reinforcement learning method~\cite{wang2020deep} and a pipeline of Siamese networks~\cite{chen2021celltrack,panteli2020siamese}.
All of these deep learning approaches predict frame-to-frame cell associations rather than providing a global tracking solution.
In contrast, for the first time (to the best of our knowledge), we present a graph neural network (GNN) framework for the tracking of multiple cell instances (represented by the graph nodes and associated by its edges) in high-throughput microscopy sequences. 
This approach allows us to simultaneously extract entire cell tracks for the construction of complete lineage trees.

A GNN is a neural network that operates on data structured as a graph. It is designed to capture graph dependencies via message passing between its nodes. Specifically, node feature vectors are computed by recursively aggregating and transforming feature vectors of their neighbors~\cite{gilmer2017neural}. The number of message passing neural network (MPNN) blocks determines the extent of the long range interactions of nodes and edges in detached frames.
Many GNN variants with different neighborhood aggregation schemes have been proposed and applied to node classification, including Graph Convolutional Network (GCN)~\cite{kipf2016semi} and Graph Attention Network (GAT)~\cite{velivckovic2017graph}. The latter is an extension of the GCN in which the weighting of neighboring nodes is inferred by an attention mechanism. In a very recent framework called Pathfinder Discovery Network Convolution (PDN-Conv), it has been proposed to use incoming edge features to weight their influence on the nodes they connect~\cite{rozemberczki2021pathfinder}. 

We extend the PDN-Conv block and design a unique MPNN paradigm based on an edge encoding network and a mutual update mechanism of node and edge features that facilitate the message passing process. Moreover, since we formulate the construction of cell trajectories by a problem of finding paths in a graph, we can address it by solving a binary edge classification problem, where `active' edges are those which connect cell instances in consecutive frames that are associated with the same biological cell. The update of the edge features is therefore an important aspect of our method.  
In~\cite{braso2020learning,weng2020gnn3dmot} GNNs have been used with different message passing schemes for vehicle and pedestrian tracking.  
Yet, cell tracking poses different challenges than tracking cars or people. 
Cell dynamics appear completely random, as there are no paths or roads neither common motion directions or speed. 
Moreover, cells within the same sequence may look very similar, yet they change their appearance along the sequence and even divide. Furthermore, in many cases, due to frequent mitotic events (cell divisions), cell population increases very rapidly such that frames, mainly toward the end of the sequence, become extremely dense and cluttered.     
There might be hundreds of cell instances in a frame, and therefore, frequent cell overlap and occlusions~\cite{eom2018phase,neumann2010phenotypic}.

Cell overlap, visual similarity, and gradual change in appearance often render manually selected features such as position, intensity, and shape insufficient for accurate tracking.
To address some of these difficulties, Arbelle et al. exploited the Kalman filter to predict changes in cell shapes in addition to their estimated pose and motion~\cite{arbelle2018probabilistic}. Some CNN-based methods learn cell features implicitly but mainly for the purpose of cell segmentation and separation of nearby cell instances.     
In this work, we use a deep metric learning (DML) technique to learn discriminative cell instance features. Specifically, we utilize the multi-similarity loss as proposed by~\cite{wang2019multi} to differentiate between instances of different cells and to associate instances of the same biological cells. The DML features of each cell instance are the input to the proposed GNN. 

Fig.~\ref{fig:model_diagram} presents the outline of our end-to-end cell tracking framework. The input to our framework is composed of a time-lapse microscopy sequence of living cells and their annotations. The outputs are the cell trajectories and lineage trees. The feature embedding networks, the GNN, and the edge classifier are connected such that the classifier loss backpropagates throughout the entire network compound.  
Our method is based on several key steps and contributions:
\textbf{1)} We address cell tracking by the simultaneous construction of all cell trajectories in the entire sequence.
\textbf{2)} We are the first to use deep metric learning to learn features that distinguish between instances of different biological cells and to assemble same cell instances.
\textbf{3)} We represent cell instances and their features as nodes in a direct graph and connect them by edges which represent their potential associations in consecutive frames. 
\textbf{4)} We are the first to apply a GNN and a message passing mechanism as a solution to the cell tracking task. Specifically, we look for paths in a graph that models the studied microscopy sequence, where each path represents a cell trajectory. 
\textbf{5)} We address cell tracking as an edge classification problem for identifying active and non-active edges. We, therefore, focus on edge features and introduce a new GNN block that enables mutual node and edge feature update. 

We demonstrate our cell tracking framework by applying it to a variety of datasets, both 2D and 3D of different cell types, imaging conditions, and experimental setups. We show that our framework outperforms current state-of-the-art methods on most of the evaluated datasets.
Specifically, we perform cell tracking on the challenging C2C12 dataset~\cite{eom2018phase} and show that our method outperforms five other competing methods and achieves state-of-the-art results. We also competed in the Cell Tracking Challenge (CTC)~\cite{10.1093/bioinformatics/btu080,ulman2017objective} as BGU-IL~$(5)$. The table published on the CTC website shows that our method was ranked first for three different datasets and second for another one.  

\section{Method}
Addressing the problem of cell tracking via GNN enables a simultaneous construction of all tracks in a sequence. Here, we present the method's building blocks and our contributions. The reader is also referred to Fig.~\ref{fig:model_diagram} for a visualization of the method outline. We consider cell tracking as a global problem applied to the entire sequence at once. This tracking model is formulated in Section~\ref{subsec:prob_form}. The representation of the entire microscopy sequence as a graph, where cell's instances are encoded by its nodes and their associations by its edges, is introduced in Section~\ref{subsec:graph_form}. This representation allows us to address cell tracking as an edge classification problem. 
In~Section~\ref{subsec:feat_extrac} we discuss cell instance feature extraction using deep metric learning.
In Section~\ref{subsec:GNN} we present the proposed GNN blocks which are designed to facilitate a message passing strategy and introduce our innovative mechanism for mutual node and edge feature update. The edge classifier and the corresponding loss function are presented in  Section~\ref{subSec:calss_train}. Finally, in Sections~\ref{subSec:inference} and~\ref{subSec:mitosis} we discuss our approach which utilizes the proposed representation for inference and mitosis detection, respectively.

\subsection{Cell Tracking Problem Formulation}\label{subsec:prob_form}
The input to our framework is a sequence of frames $\{I_t\}_{t=1}^T$ and its corresponding sequence of label maps
$\{{\mathcal{L}}_t\}_{t=1}^T,$ where $T$ is the length of the sequence. Each frame in the sequence  $I_t \colon \Omega \to \mathbb{R}+$ is a gray-scale image presenting $K_t$ cell instances, where $\Omega$ denotes a 2D or 3D image domain. The associated label map $\mathcal{L}_t \colon \Omega \to \{l^0,l^1,\ldots, l^{K_t}\}$ partitions $\Omega$ into $K_t$ connected components, each corresponds to an individual cell, and a background which is labeled by $l^0=0$ regardless of the frame number. We note that the cell labeling in each frame is arbitrary and the number of cell instances $K_t$ may vary from frame to frame due to entrance/exit of cells to/from the field of view and mitotic events.
 We further define by $s_t^k = \{\forall {\rho} \in \Omega| \mathcal{L}_t(\rho)= l^k\}$ the set of pixels/voxels $\rho$ that are associated with the $k-$th cell in frame $I_t.$ Note that $s_t^k$ can either include all the pixels/voxels that belong to a particular cell or a representative subset, depending whether $\mathcal{L}_t$ is the segmentation map of the cells in the frame or represents a set of $K_t$ markers at approximately the cell centers. 
 
Let $N$ denote the total number of biological cells that were depicted in the entire sequence, and let $c_t^n = \{s^n_t, t\}$ denote a cell object which represents an instance of the $n-$th cell in frame $t.$ We further denote by $t^n_{\mbox{\scriptsize{init}}}$ and $t^n_{\mbox{\scriptsize{fin}}}$ the first and the last time points (respectively) in which a cell is depicted in a sequence $n$. 
Note that $1\leq t_{\mbox{\scriptsize{init}}} \leq t_{\mbox{\scriptsize{fin}}} \leq T.$ To avoid confusion between the indices of the actual cells and their instances in the frame sequence, we define a cell function $\psi$ which, given a cell instance, returns the index of its corresponding biological cell. Note that if $\psi(c_t^k) = \psi(c_{t+1}^l)$ than $c_t^k$ and $c_{t+1}^l$ are instances of the same cell. 
We aim to compose a set of $N$ trajectories $\{\mathcal{T}_n\}_{n=1}^N,$ where $\mathcal{T}_n=\{c_{t_{\mbox{\scriptsize{init}}}}^n, 
 \ldots, c_{t_{\mbox{\scriptsize{fin}}}}^n\}$ is the maximal-length sequence of associated cell instances in consecutive frames that correspond to the same biological cell. Here, we set $c_t^n \equiv c_t^k$ for different values of $k$ along the sequence if $\psi(c_t^k) = n.$
 
 We assume that cells are depicted continuously; i.e., $t_{i+1} =t_{i}+1.$ Therefore, if a particular cell disappears due to an occlusion or a temporary exit from the field of view, its reappearance initiates a new track and it gets a new cell index.  To identify mitosis we define a parent function $P\colon \{1,\ldots, N\} \to \{1,\ldots, N\}$ such that $P(n) = n'$  when $n'$ is the parent of cell $n$ and $P(n) = 0$ if the cell's appearance is not a result of a mitotic event. For convenience we set $n'<n$. Note that a cell cannot be a parent of itself, therefore $n' \neq n.$ 
 A complete trajectory object of the $n-$th biological cell is defined as follows, where ${\boldsymbol{\mathcal{T}_n}}= \{{\mathcal{T}}_n, P(n),t^n_{\mbox{\scriptsize{init}}},
 t^n_{\mbox{\scriptsize{fin}}}\}.$
The cell tracking task goal is to find the set $\{{\boldsymbol{\mathcal{T}_1}}, ..., {\boldsymbol{\mathcal{T}_N}}\}$ that best explains the observations.

\subsection{Graph Formulation}\label{subsec:graph_form}
We use a direct, acyclic graph to model cell-to-cell associations in microscopy sequences.
Let $\mathcal{G}=(\mathcal{V},\mathcal{E})$ define a graph represented by its vertices (nodes) $\mathcal{V}$ and edges (links) $\mathcal{E}.$ Let $M = \sum_{t=1}^T K_t$ represent all cell instances in the entire frame sequence.
A graph representation of cells and their associations is composed of $|\mathcal{V}|=M$ vertices, where each node $\nu_i \in \mathcal{V},$ $i=1, \ldots, M$ represents a single cell instance $c_{t= \tau}^k.$ For convenience, we can set $i = \sum_{t=1}^{\tau-1}K_t + k.$

An edge $e_{i,j} \in \mathcal{E}$ represents a potential association between a pair of vertices $\nu_i, \nu_j$, representing cell instances in  consecutive frames. To reduce the number of edges, we connect a pair of cells only if their spatial Euclidean distance is within a neighborhood region, which is calculated based on the training set
as detailed in Section~\ref{subsec:graph_const} in the Appendix.

We address cell tracking as an edge classification problem. The desired output are labeled sets of edges
defined by an edge function $Y\colon \mathcal{E} \to \{0,1\}.$ Let $\nu_i, \nu_j$ represent cell instances denoted by $c_t^k$ and $c_{t+1}^l$, respectively. 
\begin{equation}
    \centering
    Y(e_{i,j})= y_{i,j} =
    \begin{dcases}
        1,& \text{if $\psi(c_t^k) = \psi(c_{t+1}^l$)}\\
        0,              & \text{otherwise}
    \end{dcases}    
\end{equation}
Accurate prediction of $Y(e_{i,j})$ for the complete set of graph edges provides the entire cell lineage associated with the observed microscopy sequence. 
A cell trajectory $\mathcal{T}_n$ can be either defined  by a sequence of cell instances represented by the graph's vertices $\{\nu_{i_1},\nu_{i_2}, \ldots \nu_{i_n} \}$, or by a sequence of edges $\mathbf{e}_n = \{ e_{i_1i_2}, \ldots, e_{i_{n-1}i_n} \},$ that connect cell instances in consecutive frames, where $\{ e_{ij} \in \mathbf{e}_n ~|~ Y(e_{ij}) =1\}.$ 

We assume that a non-dividing cell instance has at most a single successor while a cell that undergoes mitosis may have two successors and even more (in rare occurrences). Ideally, if two (or more) different nodes in a frame $\nu_j \neq \nu_{j'}$ are connected to the same node $\nu_i$ in a previous frame,
i.e.; $Y(e_{i,j}) =Y(e_{i,j'}) =1,$ then we can assume that  $P(\psi(c_{t+1}^l)) = P(\psi(c_{t+1}^{l'})) = \psi(c_t^k)$ and detect a mitosis event. In practice, often the visual features of daughter cells differ from those of their parent and an additional process is required to identify and validate parent-daughter relations.
The complete representation of the proposed graph is defined by the following attribute matrices: i) A node feature matrix $\textbf{X}\in \mathbb{R}^{|\mathcal{V}|\times d_\mathcal{V}}$ with $d_\mathcal{V}$ features per node.
ii) A graph connectivity matrix $\textbf{E}\in \mathbb{N}^{2\times|\mathcal{E}|}$ which represents all possible linked cell indices from source to target. iii) An edge feature matrix $\textbf{Z}\in \mathbb{R}^{|\mathcal{E}|\times d_\mathcal{E}}$, where each row in $\textbf{Z}$ consists of $d_\mathcal{E}$ features of an edge $e_{ij}$ in the graph. We aim to predict $\hat{\textbf{Y}}\in \mathbb{R}_{[0,1]}^{|\mathcal{E}|\times 1}$ that represents the probability to represent an actual cell association. Next, we present the initial graph embedding.

\subsection{Feature Extraction}\label{subsec:feat_extrac}
The success of cell tracking algorithms depends on correct associations of instances of the same biological cells.  
We consider instances of the same cell as members of the same class. Altogether, we have $N$ mutually exclusive classes.
\noindent\textbf{Deep Metric Learning (DML) Features.}\label{subsubsec:DML}\quad
We use \textit{deep metric learning} to learn cell feature embeddings that allow us to assemble instances of the same biological cells and distinguish between different ones.
For this purpose, we use the cell segmentation maps or marker annotations to crop each frame into sub-images of all cell instances.
Following~\cite{wang2019multi} we use a \textit{hard mining} strategy and a \textit{multi-similarity loss} function
to train a ResNet network~\cite{he2016deep} to predict such embeddings. 
Specifically, we generate batches of cell sub-images, where each is composed of $m$ same-class instances from $\kappa$ classes. Since the cell's appearance gradually changes during the sequence we perform the \textit{m-per-class} sampling~\cite{musgrave2020metric} using temporally adjacent frames.  To calculate the loss, we consider pairs of within-class and between-class cell instances which form the positive and negative examples, respectively. To conduct the hard mining, an affinity matrix $\mathbb{A} \in \mathbb{R}^{m\kappa \times m\kappa}$ is constructed based on the cosine similarity function applied to the `learned' feature vectors of each of the pairs. A column $i$ in $\mathbb{A}$ presents the proximity of all the batch instances to the $i-$th instance. An instance $j$ forms a \textit{hard negative} example with $i$ if $i$ and $j$ are not within the same class and $j$ is more similar to $i$ than another instance $k$ that form a positive example with it. In the same manner, an instance $l$ forms a \textit{hard positive} example with $i$ if $i$ and $l$ are within the same class and $l$ is less similar to $i$ than another instance $r$ that form a negative example with it.
The multi-similarity loss is a measure of the \textit{hard positive} and \textit{hard negative} examples. When there are no such examples the embedded feature vectors are adequately clustered. 
The node feature vectors extracted using DML approach are denoted in a matrix form as follows: ${\mathbf V}_{\mbox{\scriptsize{DML}}} \in \mathbb{R}^{|\mathcal{V}| \times d_{\mbox{\scriptsize DML}}}, $ 
where $d_{\mbox{\scriptsize DML}}$ denotes the number of DML features.  

\noindent\textbf{Spatio-temporal Features.}\label{subsubsec:ST}\quad
While the learned feature vectors serve to distinguish between cell instances based on their visual appearance alone they do not account for temporal and global spatial features that also characterize the cells. These include the coordinates of the cell center, its frame number, and intensity statistics (minimum, maximum, and average). In the case where we have an instance segmentation mask, the cell's area, the minor and major axes of a bounding ellipse, and bounding-box coordinates are also considered. We denote by $\textbf{V}_{\mbox{\scriptsize{ST}}} \in \mathbb{R}^{|\mathcal{V}| \times d_{\mbox{\scriptsize ST}}}$ the 
spatio-temporal (ST) feature matrix, which is composed of 
the $d_{\mbox{\scriptsize ST}}$-dimensional feature vectors of all nodes. 

\noindent\textbf{Initial Edge and Node Features.}\label{subsubsec:InitFeatures}\quad
The complete feature matrix of all nodes in the graph includes both the learned and the spatio-temporal features and is denoted by $\textbf{V}_{\mathcal{V}} \in \mathbb{R}^{|\mathcal{V}| \times d_{\mathcal{V}}},$ where, $d_{\mathcal{V}}=d_{\mbox{\scriptsize ST}}+ d_{\mbox{\scriptsize DML}}.$  
It is generated by a concatenation of $\textbf{V}_{\mbox{\scriptsize{DML}}}$ and $\textbf{V}_{\mbox{\scriptsize{ST}}}.$ 
Having $\textbf{V}_{\mathcal{V}}$ we construct the initial edge feature matrix $\textbf{V}_{\mathcal{E}}$ using the \textit{distance $\&$ similarity} operation defined in Eq.~\ref{eq:DS_block}. 
Since $\textbf{V}_{\mbox{\scriptsize{DML}}}$ and $\textbf{V}_{\mbox{\scriptsize{ST}}}$ are from different sources and are in different scales, we homogenize them and reduce the complete feature vector dimension via mapping by multi-layer perceptron (MLP) networks. These MLPs are connected to the proposed GNN (see Section~\ref{subsec:GNN}) in an end-to-end manner.  
We denote the initial node feature vector of a vertex $\nu_i$ by $\mathbf{x}^{(0)}_i, $ where $\mathbf{x}^{(0)}_i$ is the $i$-th row in $\mathbf{X}^{(0)}.$

\begin{figure}[t]
    \centering 
    \subfigure[GNN]{\label{fig:GNN}\includegraphics[width=95mm]{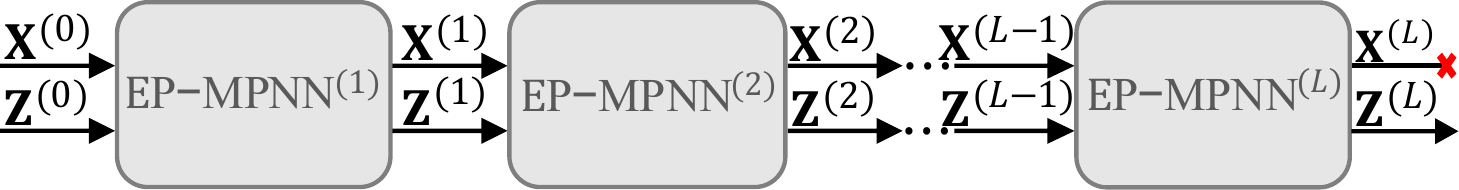}}\\
    \subfigure[EP-MPNN]{\label{fig:EP_MPNN}\centering\includegraphics[width=50mm]{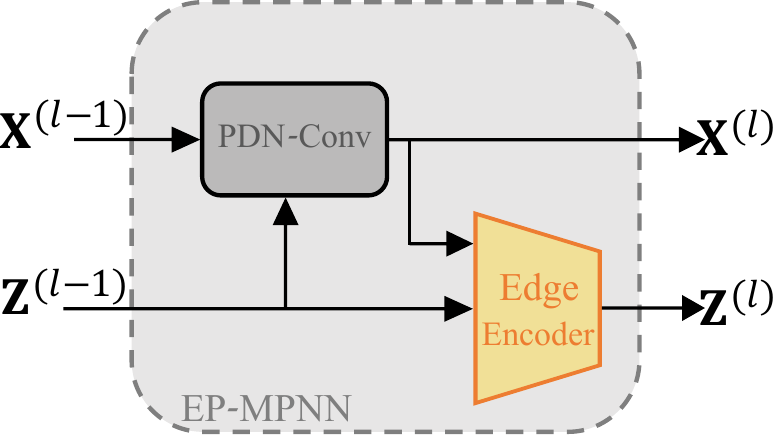}} \hskip 1.0cm 
    \subfigure[Edge Encoder]{\label{fig:edge_encoder}\centering\includegraphics[width=30mm]{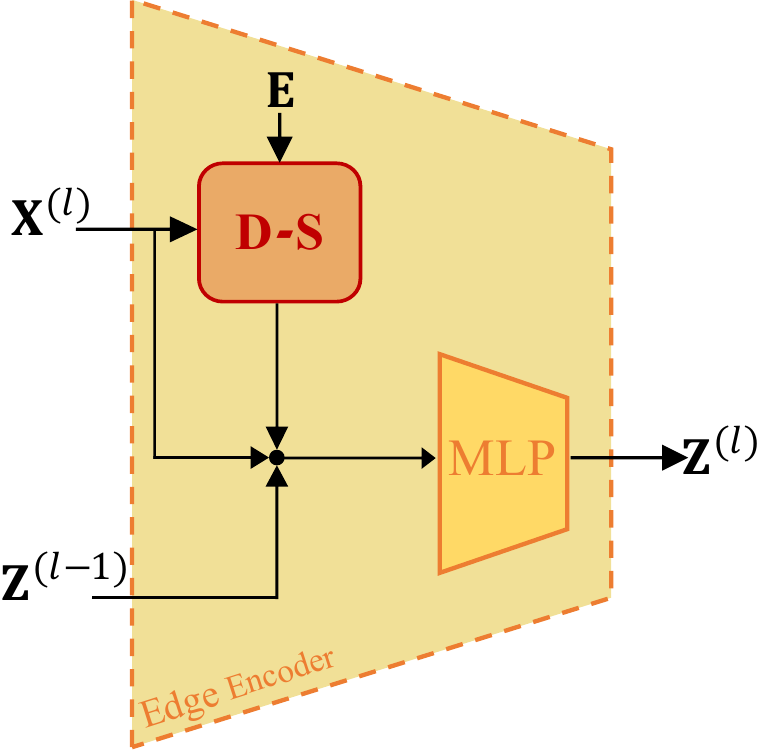}}
    
    \caption{
    \textbf{(a) A Graph Neural Network (GNN).} The GNN is composed of $L$ EP-MPNN blocks where $L$ determines the message passing extent. The $l$-th EP-MPNN block
    updates the nodes and edge features, i.e., $\mathbf{X}^{(l-1)} \to \mathbf{X}^{(l)} $
    and $\mathbf{Z}^{(l-1)} \to \mathbf{Z}^{(l)}.$
    \textbf{(b) Edge-oriented Pathfinder - Message Passing Network (EP-MPNN) Layer}. The basic layer in the graph neural network step comprises a PDN-Conv and an edge encoder. The PDN-Conv updates the node feature vectors based on their current values and an edge attention model.
    \textbf{(c) Edge encoder.} Updates the edge feature vectors. Its \textbf{\textit{D-S}} block calculates the distance and similarity between the feature vectors of each pair of nodes by an edge. The output of the \textbf{\textit{D-S}} block along with the nodes and the current edge features compose the input of an MLP which outputs the new feature vectors of the edges. 
    The concatenation (denoted by '$\otimes$') of the \textbf{\textit{D-S}} block's output along with the current node and edge features formulated in Eq.~\ref{eq:PDN_EDGE}, is the input for an MLP which is trained to learn a new edges representation
    }
\label{fig:Plot_blocks}
\end{figure}

\subsection{Graph Neural Network}\label{subsec:GNN}
The core of the proposed cell tracking framework is the graph neural network (GNN) illustrated in Fig.~\ref{fig:GNN}. Exploiting the GNN model and the message passing paradigm allows us to simultaneously trace entire cells tracks rather than locally associate cell instances in a frame-by-frame manner.
One of our main contributions is the graph message passing block presented in Fig.~\ref{fig:EP_MPNN} called the Edge-oriented Pathfinder Message Passing Neural Network (EP-MPNN). Specifically, we extend the MPNN block presented in~\cite{rozemberczki2021pathfinder} by introducing an edge encoder, thus enabling an interplay between the edge and node feature update simultaneously with an edge-attention mechanism.  

The GNN is composed of $L$ EP-MPNN blocks where $L$ determines the message passing extent. In other words, the associations of cell instances in consecutive frames are affected by the respective connections along the sequence up to $L$ frames away.
The input to the $l+1$-th EP-MPNN block (which is, in fact, the output of the $l$-th EP-MPNN block) is composed of the updated node and edge features, denoted by  $\mathbf{X}^{(l)}=\{x^{(l)}_{i}\}_{\nu_i\in\mathcal{V}}$ and $\textbf{Z}^{(l)}=\{z^{(l)}_{ij}\}_{ e_{ij}\in\mathcal{E}},$
respectively, where $l = 0, \ldots, L.$
The nodes are updated using the pathfinder discovery network convolution (PDN-Conv)~\cite{rozemberczki2021pathfinder} which is one of the two EP-MPNN components. The other component is the \textit{edge encoder} (illustrated in Fig.~\ref{fig:edge_encoder}) which is trained to embed the edge features based on the node features. 
We introduce the edge encoder in the following.

\noindent\textbf{Node Feature Update.}\label{subsubsec:Nodeupdate}\quad
The features of each node $\nu_i$ are updated based on the weights of the incoming edges. These weights are learned using an attention mechanism.   Let $f^{\mbox{\scriptsize{PDN}}}_{\mbox{\scriptsize{edge}}} \colon \mathbb{R}^{d_\mathcal{E}}\to \mathbb{R}$ define a function implemented by an MLP that is trained to output scalars which represent the weights of the edges, given their current features. Let $f^{\mbox{\scriptsize{PDN}}}_{\mbox{\scriptsize{node}}}\colon \mathbb{R}^{d_\mathcal{V}}\to \mathbb{R}^{d_\mathcal{V}}$ define a vector function implemented by an MLP that is trained to output updated feature vectors given the current ones.
The updated feature vector of a node $\nu_i$ is obtained by a weighted sum of its own and its neighboring nodes, as follows: 
\begin{equation}\label{eq:PDN_NODE}
    \centering
    \mathbf{x}^{(l)}_i= \sum_{j\in \mathcal{N}(i)\cup \{i\} } \overbrace{f^{\mbox{\scriptsize{PDN}}}_{\mbox{\scriptsize{edge}},l} (\mathbf{z}^{(l)}_{j,i})}^{\omega^l_{ji}}  \overbrace{f^{\mbox{\scriptsize{PDN}}}_{\mbox{\scriptsize{node}},l} (\mathbf{x}^{(l-1)}_{j})}^{\tilde{\mathbf{x}}^{(l-1)}_{j}},
\end{equation}
where $\mathcal{N}(i)$ denotes the neighbors of $\nu_i;$ i.e., all the nodes $\nu_j$ for which there exist $e_{j,i} \in \mathcal{E}.$
Note that $f^{\mbox{\scriptsize{PDN}}}_{\mbox{\scriptsize{edge}},l}$ and $f^{\mbox{\scriptsize{PDN}}}_{\mbox{\scriptsize{node}},l}$ are trained separately for each block.
Eq.~\ref{eq:PDN_NODE} can be interpreted as attention through edges, 
where, $\omega^l_{ji} =f^{\mbox{\scriptsize{PDN}}}_{\scriptsize{edge},l}(\mathbf{z}^{(l)}_{j,i})$ is the predicted attention parameter of an edge $e_{j,i}$ ($\omega^l_{ii}=1$) and $\tilde{\mathbf{x}}^{(l-1)}_{j} =f^{\mbox{\scriptsize{PDN}}}_{\mbox{\scriptsize{node}},l} (\mathbf{x}^{(l-1)}_{j})$ is the \textit{mapped} feature vector of a node $\nu_j$ in the $l$-th EP-MPNN. 

\noindent\textbf{Edge Feature Update.}\label{subsubsec:Edgeupdate}\quad
The main contribution of the proposed GNN framework is a mechanism for edge feature update that enhances the message passing process. Unlike the GNN presented in~\cite{rozemberczki2021pathfinder} here, the edge and node features are alternately updated. We denote by $\textit{\textbf{D-S}}$ a function that returns the distance \& similarity vector of two feature vectors of connected nodes as follows: 
\begin{equation}\label{eq:DS_block}
    \centering
    \textit{\textbf{D-S}}(\mathbf{v}_i, \mathbf{v}_j) = \big{[}|v_i^1-v_j^1|, \ldots,|v_i^{d_{\nu}}-v_j^{d_{\nu}}|,
    \frac{\mathbf{v}_i \cdot \mathbf{v}_j}{\lVert\mathbf{v}_i\lVert\lVert\mathbf{v}_j\lVert}\big{]}
\end{equation}
which is a concatenation of the absolute values of the differences between corresponding elements in $\mathbf{v}_i$ and  $\mathbf{v}_j$ and their cosine similarity $\frac{\mathbf{v}_i \cdot \mathbf{v}_j}{\lVert\mathbf{v}_i\lVert\lVert\mathbf{v}_j\lVert}$. 
In the $l$-th block $\mathbf{v}_i$ = $\mathbf{x}_i^{(l)}$ and $\mathbf{v}_j$ = $\mathbf{x}_j^{(l)}.$ In the initial phase, when the input to the first GNN block is formed, $\mathbf{v}_i$ and $\mathbf{v}_j$ are the $i$-th and the $j$-th rows in  $\textbf{V}_{\mathcal{V}},$ respectively. The construction of the initial feature matrix $\textbf{V}_{\mathcal{V}}$ is described in Section~\ref{subsubsec:InitFeatures}.

The edge update function $f^{\mbox{\scriptsize{EE}}}_{\mbox{\scriptsize{edge}},l}$ (implemented as an MLP) returns the updated features of each edge $e_{i,j}$ given the concatenation of the current edge features, the updated feature vectors of the nodes it connects, and the output of the $\textit{\textbf{D-S}}$ block applied to these nodes. 
Formally,
\begin{equation}\label{eq:PDN_EDGE}
    \mathbf{z}^{(l)}_{ij} = f^{\mbox{\scriptsize{EE}}}_{\mbox{\scriptsize{edge}},l} ([ \mathbf{z}^{(l-1)}_{ij},  \mathbf{x}^{(l)}_i,  \mathbf{x}^{(l)}_j,\textit{\textbf{D-S}}( \mathbf{x}^{(l)}_i,  \mathbf{x}^{(l)}_j)])
\end{equation}
The simultaneous edge and node feature update facilitates the message passing mechanism. Let $\nu_i$ be a node in frame $t$ and let $\mathcal{N}(i)$ define the set of nodes connected to it in frame $t-1.$ Consider a node $\nu_j \in \mathcal{N}(i).$ In the first GNN block ($l=1$) the features of an edge $e_{j,i}$ are updated based on the features of $\nu_i$ and $\nu_j.$ In the same block $\nu_j$ is updated based on the features of its neighbors $\mathcal{N}(j)$ that are nodes representing cell instances in frame $t-2.$ The update is determined by the weights of the edges connecting them. In the next GNN block ($l=2$) the features of an edge $e_{j,i}$ are again updated based on the features of $\nu_i$ and $\nu_j.$ However, this time the features of a node $\nu_j$ are already influenced by its neighbors. Therefore, when the features of $\nu_i$ are updated they are already affected by the neighbors of its neighbors which are cell instances that are two frames away. In the same manner, in the $L$-th block, cell instances and their first-order connections are influenced by cell-to-cell associations and higher-order connections along $L$ frames. 

\subsection{Classifier and Training} \label{subSec:calss_train}
The output edge feature vectors are the inputs to the edge classifier network. The classifier is an MLP with three linear layers each is followed by a ReLU activation, when a Sigmoid function is applied to the output layer. The output is a vector $\hat{\textbf{Y}}\in \mathbb{R}_{[0,1]}^{|\mathcal{E}|\times 1}$ that represents the probability for each edge to be active ($=1$) or not ($=0$). We use the ground-truth (GT) edge activation vector $\textbf{Y}$ to train the model. Since most of the edges in the graph dataset are not active (i.e., do not link nodes), our data are highly imbalanced. Therefore, we use a weighted cross-entropy loss function with adaptive weights which are determined by the average number of neighbors in a batch, i.e., $\big(\frac{1}{|\mathcal{N}|}, \frac{|\mathcal{N}|- 1}{|\mathcal{N}|}\big)$. Note that since the size of the neighborhood region remains fixed throughout the sequence, the number of neighbors increases as the frames become denser.

\subsection{Cell Tracking Inference} \label{subSec:inference}
The output of the proposed deep learning framework is a probability matrix which identifies active edges. It is used together with the connection matrix $\mathbf{E}$ to construct candidates for cell trajectories as described in Section~\ref{subsec:graph_form}.
Specifically, predictions of cell tracks are obtained at the inference phase in the form of a directed graph with soft edge weights (the output of a sigmoid). The edges in the graph represent only outgoing, potential associations between consecutive frames. The soft weights (association probabilities) allow us to partition the graph edges into active and non-active. Considering the outgoing/incoming edges of a specific node, there could be one of the following outcomes: $1)$ All outgoing/incoming edges are non-active - which  may indicate end/beginning of a track. 
$2)$ Only one outgoing edge is active. $3)$ Two or more outgoing edges are active which may  indicate mitosis (cell division). $4)$ More than a single incoming edge is active - i.e., when different cell instances are associated to the same cell instance. 
Above $99\%$ of incoming/outgoing edges conflicts are avoided thanks to the proposed training scheme. 
We note that the network is implicitly trained to prefer bijection cell associations thanks to the attention-based mechanism of the GNN blocks and the weighted loss function (see Section~\ref{subSec:calss_train}). 
Nevertheless, to ensure one-to-two mapping at most (case 3), in case that the association probabilities of more than two outgoing edges are higher than 0.5 (extremely rare events) only the top-2 are considered as active.
 In addition, to ensure injective mapping (case $4$), only the incoming edge with the highest association probability (as long as it is higher than 0.5) is considered active. This obviously feasible ad-hoc strategy ensures a single path to each cell. 

\subsection{Mitosis Detection} \label{subSec:mitosis}
 Since daughter cells usually have different visual features than their parent, it is not very frequent for a node to have two active outgoing edges. In most mitotic events the parent track terminates whereas two new tracks initiate. To associate pairs of daughter cells to their parents we consider the detected tracks of all cells. We then look for triplets of trajectories $(\mathbf{\mathcal{T}}_k,\mathbf{\mathcal{T}}_l,\mathbf{\mathcal{T}}_m)$ where $k,l,m \in \{1, \ldots, N\}$ such that  $t^{k}_{\mbox{\scriptsize{init}}} =t^{l}_{\mbox{\scriptsize{init}}} =t^{m}_{\mbox{\scriptsize{fin}}}+1.$   
If the spatial coordinates of $c_{t_{\mbox{\scriptsize{init}}}}^k,c_{t_{\mbox{\scriptsize{init}}}}^l$ and 
$c_{t_{\mbox{\scriptsize{fin}}}}^m$ are within the same neighborhood region, then we set $P(k) = P(l) = m.$

\begin{figure*}[t]
\centerline{\includegraphics[scale=0.227,trim={0cm 0.0cm 0 0cm},clip]{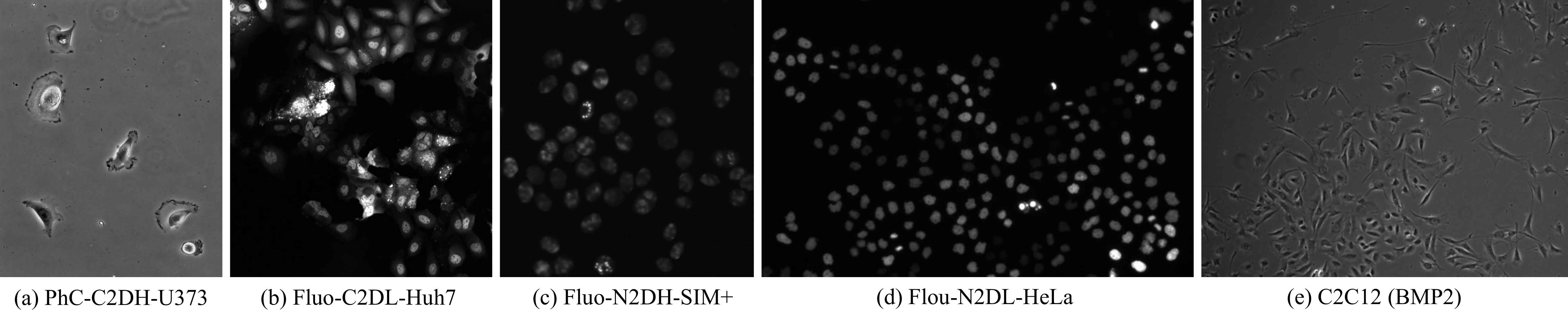}}

  \caption{Example frames from five evaluated 2D datasets. Note the different appearance of the cells and the entire frames
  \label{fig:dataset_visualization}}

\end{figure*}

\section{Experiments} \label{Sec:Exp}

We evaluated our model on six datasets from two different sources: an open dataset~\cite{eom2018phase} we call C2C12 and five (2D and 3D) datasets from the Cell Tracking Challenge (CTC)~\cite{10.1093/bioinformatics/btu080,ulman2017objective}.
The C2C12 dataset is composed of $16$ sequences acquired with four different cell growth factors. 
Representative frames are presented in Fig.~\ref{fig:dataset_visualization} for each of the 2D data sequences. Implementation details and further experiments including qualitative results (figures and videos) and ablation studies are provided in Sections~\ref{sec:imp_details} and~\ref{sec:exp} (respectively) of the appendix.

\begin{table*}[t]
\caption{\textbf{C2C12 cell tracking results~\cite{eom2018phase}.} 
      A comparison of our association accuracy (left) and target effectiveness (right) scores to the scores obtained by five other methods. The top five rows present results for the original frame rate (best appears in \textbf{bold}) and the bottom two rows refer to the lower ($\times5$) frame rate setup (best is \underline{underlined})}
  \centering
    \renewcommand{\arraystretch}{0.9}
\setlength{\tabcolsep}{4pt}
\scalebox{0.81}{
    \begin{tabular}{lccccccccccc}
    \toprule
          & \multicolumn{5}{c}{\textbf{Association Accuracy}} &       & \multicolumn{5}{c}{\textbf{Target Effectiveness}} \\
\cmidrule{2-6}\cmidrule{8-12}    \textbf{Method} & control & FGF2  & BMP2  & \begin{tabular}[c]{@{}l@{}}FGF2+ \\BMP2\end{tabular} & Avg.  &       & control & FGF2  & BMP2  & \begin{tabular}[c]{@{}l@{}}FGF2+ \\BMP2\end{tabular} & Avg. \\
    \midrule
    AGC~\cite{bensch2015cell} & 0.604 & 0.499 & 0.801 & 0.689 & 0.648 &       & 0.543 & 0.448 & 0.621 & 0.465 & 0.519 \\
    ST-GDA~\cite{bise2011reliable}  & 0.826 & 0.775 & 0.855 & 0.942 & 0.843 &       & 0.733 & 0.710 & 0.788 & 0.633 & 0.716 \\
    BFP~\cite{nishimura2020weakly} & 0.955 & 0.926 & 0.982 & 0.976 & 0.960 &       & 0.869 & 0.806 & 0.970 & 0.911 & 0.881 \\
    MPM~\cite{Hayashida_2020_CVPR}   & 0.947 & 0.952 & \textbf{0.991} & \textbf{0.987} & 0.969 &       & 0.803 & 0.829 & 0.958 & 0.911 & 0.875 \\
    Ours  & \textbf{0.981} & \textbf{0.973} & \textbf{0.991} & 0.986 & \textbf{0.983} &       & \textbf{0.894} & \textbf{0.843} & \textbf{0.976} & \textbf{0.914} & \textbf{0.907} \\
    \midrule
          
          
    CDM($\times5$)~\cite{hayashida2019cell} & 0.883 & 0.894 & 0.971 & 0.951 & 0.927 &       & 0.832 & 0.813 & 0.958 & 0.895 & 0.875 \\
    Ours ($\times5$) & \underline{0.958} & \underline{0.942} & \underline{0.988} & \underline{0.970} & \underline{0.964} &       & \underline{0.905} & \underline{0.852} & \underline{0.981} & \underline{0.919} & \underline{0.914} \\
    \bottomrule
    \end{tabular}%
    }\label{tab:C2C12_results}%
\end{table*}%
\subsection{C2C12 Myoblast Datasets} \label{subSec:data}
In the following experiments, we focus on C2C12~\cite{eom2018phase}, which is considered a challenging cell tracking dataset (see trajectories in Fig.~\ref{fig:model_diagram}). It is composed of high-throughput time-lapse sequences of C2C12 mouse myoblast cells acquired by phase-contrast microscopy under four growth factor conditions, including with fibroblast growth factor 2 (FGF2), bone morphogenetic protein 2 (BMP2), FGF2 + BMP2, and control (no growth factor). For each condition, four image sequences were captured, each consists of $780$ frames ($12$ frames-per-hour, $1392 \times 1040$ pixels).
The GT annotation includes a list of coordinates of the cell instances' centers along with the corresponding indices of the depicted biological cells.
There is a complete cell annotation for only one sequence, where three cells and their lineage trees were marked in the others. 

\noindent\textbf{Experimental Setting and Evaluated Metrics.}\quad
To quantify the tracking performance for the C2C12 dataset, we used the association accuracy (AA) and the target effectiveness (TE) measures as suggested by~\cite{Hayashida_2020_CVPR,nishimura2020weakly}. All comparisons were made with respect to human-annotated target cells and their frame-to-frame associations, which are considered the GT.  The AA measure is defined by the ratio between the number of true positive associations and the total number of GT associations. The TE measure~\cite{blackman1986multiple} considers the number of cell instances correctly associated within a track with respect to the total number of GT cell instances associated within that track. 
Training-test split was as in \cite{Hayashida_2020_CVPR} where the test sequences of the C2C12 were annotated by the nnU-Net network \cite{isensee2021nnu}. 

\noindent\textbf{C2C12 Tracking Results.}\quad
The AA and the TE scores are reported in Table~\ref{tab:C2C12_results} for the four different types of C2C12 growth factor conditions. The top part of the table compares the performances of our framework with respect to the reported scores of four other methods: AGC~\cite{bensch2015cell}, ST-GDA~\cite{bise2011reliable}, BFP~\cite{nishimura2020weakly}, MPM~\cite{Hayashida_2020_CVPR}, where the latter two were considered the best performing methods for this dataset~\cite{Hayashida_2020_CVPR,nishimura2020weakly}. 
The bottom part of the table presents tracking scores for subsampled C2C12 sequences, simulating $\times 5$ lower frame rate. Our scores are compared with those obtained by the CDM method~\cite{hayashida2019cell}
which was reported to have the best performance for this setup.
Our tracking framework is shown to outperform all other methods, for almost all cell growth factor conditions in both setups, possibly thanks to the following aspects. 
Since the optimization process is performed via training through the loss functions, our method is advantageous over the classical global cell tracking approaches which are based on combinatorial algorithms such as AGC~\cite{bensch2015cell} and ST-GDA~\cite{bise2011reliable}.  
Moreover, the uniquely designed message passing scheme which considers long term cell associations is preferable over the local, greedy and therefore sub-optimal frame-by-frame association paradigms such as MPM~\cite{Hayashida_2020_CVPR}, BFP~\cite{nishimura2020weakly}, and CDM~\cite{hayashida2019cell}.


\begin{table*}[t]
  \centering
    \renewcommand{\arraystretch}{1.3}
          \caption{\textbf{Cell Tracking Challenge (CTC) results.} Our tracking scores (TRA measure~\cite{matula2015cell}) for the test sequences of five different datasets as published by the CTC organizers. The table presents the top three scores (our scores appear in bold) and our ranks with respect to the number of competing methods. The difference between our score and the best one appears in parentheses}
\setlength{\tabcolsep}{6pt}
\scalebox{0.85}{
    \begin{tabular}{c|c|c|c|c}
    \hline 
    Dataset & Our-Rank(diff) & 1st   & 2nd   & 3rd \\
    \hline 
    \hline 
        PhC-C2DH-U373 &$\textbf{1}/25$& \textbf{0.985} &0.982&0.981\\
    \hline 
    Fluo-N2DH-SIM+ &$\textbf{1}/35$& \textbf{0.978} &0.975&0.973\\
    \hline 
    Fluo-N3DH-SIM+ &$\textbf{1}/11$& \textbf{0.974} &0.972&0.967  \\
    \hline 
    Fluo-C2DL-Huh7 &$2/6(0.026)$&0.960& \textbf{0.934} &0.865\\
    \hline 
    Fluo-N2DL-HeLa &$4/34(0.004)$&0.993&0.991&0.991\\
    \hline 
    \end{tabular}
    }
  \label{tab:CTB}%
\end{table*}%

\subsection{Cell Tracking Challenge (CTC)} \label{subSec:CTC}
We competed in the CTC for five datasets~\cite{10.1093/bioinformatics/btu080,ulman2017objective} to further assess our method. 
Fig.~\ref{fig:dataset_visualization} presents representative examples for each of the datasets, demonstrating their variability. The CTC is an excellent platform for objective and unbiased comparisons of the performances of different cell tracking methods when applied to a variety of 2D and 3D microscopy sequences of different acquisition methods (e.g., fluorescence or phase-contrast microscopy) and different cell types. Each CTC dataset is split into training and test sequences where the annotations of the test sets are deliberately unavailable to the competitors. The scores and ranks for all competing methods are calculated and published by the challenge organizers\footnote{The Cell Tracking Benchmark web page is available at \url{http://celltrackingchallenge.net/latest-ctb-results/}.}.
To quantify the tracking results the CTC organizers use the TRA measure~\cite{matula2015cell}. The TRA measure is based on a comparison of the nodes and edges of acyclic oriented graphs representing cells and their associations in both the GT and the evaluated method. It computes the weighted sum of the required operations to transform the predicted cell lineage tree into the GT cell lineage tree and captures all the required information for tracking assessment, including mitotic events.
For the CTC experiments, we used a publicly available segmentation method~\cite{arbelle2019microscopy} to predict segmentation maps for the test datasets.

\noindent\textbf{Results.}\quad
In Table~\ref{tab:CTB} we present the TRA scores for five CTC datasets: PhC-C2DH-U373, Fluo-N2DH-SIM+, Fluo-N3DH-SIM+, Fluo-C2DL-Huh7 and Fluo-N2DL-HeLa. Our scores are compared to the scores of the three-best performing methods. Our ranks are presented with respect to the number of competing methods that took part in the CTC for each dataset. Our method was ranked \textbf{first} for three out of five datasets and achieved the second- and the fourth-best ranks for the two other datasets. 
Our tracking results appear under \href{http://celltrackingchallenge.net/participants/BGU-IL/}{BGU-IL~$(5)$}.


\begin{table*}[t]
  \centering
    \renewcommand{\arraystretch}{1.0}
          \caption{\textbf{Quantitative ablation study.} ST and DML stand for spatio-temporal and deep-metric-learning feature extraction, respectively. EE stands for the proposed edge encoder (Fig.~\ref{fig:edge_encoder}). MPNN is the chosen message passing scheme. The symbols \ding{51} and \ding{55} indicate included or excluded, respectively. AA and TE stand for association accuracy and target effectiveness scores}
\setlength{\tabcolsep}{6pt}
\scalebox{0.85}{
 \begin{tabular}{c|c|c|c|r|r}
    \hline
    \multicolumn{4}{c|}{Components} & \multicolumn{2}{c}{Results} \\
    \hline
    ST   & DML  & EE & MPNN  & \multicolumn{1}{c|}{AA} & \multicolumn{1}{c}{TE} \\
    \hline
    \hline
    \ding{51}     & \ding{51}     & \ding{51}     & GAT   &            0.805  &        0.353  \\
    \ding{51}     & \ding{51}     & \ding{55}     & GCN   &            0.775  &        0.271  \\
    \ding{51}     & \ding{51}     & \ding{51}     & GCN   &            0.871  &        0.359  \\
    \ding{51}     & \ding{51}     & \ding{55}     & PDN   &            0.205  &        0.109  \\
    \ding{55}    & \ding{51}     & \ding{51}     &EP-MPNN   &            0.989  &        0.960  \\
    \ding{51}     & \ding{55}     & \ding{51}     & EP-MPNN   &            0.987  &        0.978  \\
    \ding{51}  & \ding{51}  & \ding{51}  & EP-MPNN & \textbf{0.994} & \textbf{0.989} \\
    \end{tabular}
    }
  \label{table:ablation_main}
\end{table*}%

\subsection{Ablation Studies}
\label{ssec:Ablation}

We conduct ablation studies using the C2C12 dataset~\cite{eom2018phase} to demonstrate the contribution of each component of our framework to the overall cell tracking performance.
  The first three rows in Table~\ref{table:ablation_main} present a comparison of the proposed EP-MPNN to two other commonly used message passing schemes, namely, GCN~\cite{kipf2016semi} and GAT~\cite{velivckovic2017graph}. The PDN method is similar to ours but does not have the edge encoder in the GNN block. The comparisons to the PDN and to the GCN without edge encoder assessed its importance. 
  To demonstrate the significance of both the DML and the ST features we conducted two more experiments that are presented in the fifth and the sixth rows in the table.
 The DML and ST features both contribute to our tracking results. Further ablation study experiments are presented in Section~\ref{sec:ablation_cont} of the Appendix.

\section{Conclusions}\label{sec:Discussion and Conclusions}

We introduced an end-to-end deep learning framework for the simultaneous detection of complete cell trajectories in high-throughput microscopy sequences.  
This was accomplished by representing cell instances and their potential associations by nodes and edges (respectively) in a direct, acyclic graph; using deep metric learning for the extraction of distinguishing node features; 
and by using GNN and a uniquely designed message passing scheme to apply long range interactions between nodes and edges. The GNN's edge feature outputs were exploited to detect active edges that form paths in the graph, where each such path represents a cell trajectory.

Ablation study experiments presented in Section~\ref{ssec:Ablation} assess the contribution of each of the proposed method components to the tracking accuracy. State-of-the-art tracking results are shown for a variety of 2D and 3D publicly available microscopy sequences. The code is released for comparisons and future study.




\clearpage
%
%
\bibliographystyle{splncs04}
\bibliography{egbib}

\begin{thebibliography}{10}
\providecommand{\url}[1]{\texttt{#1}}
\providecommand{\urlprefix}{URL }
\providecommand{\doi}[1]{https://doi.org/#1}

\bibitem{amat2014fast}
Amat, F., Lemon, W., Mossing, D.P., McDole, K., Wan, Y., Branson, K., Myers,
  E.W., Keller, P.J.: Fast, accurate reconstruction of cell lineages from
  large-scale fluorescence microscopy data. Nature methods  \textbf{11}(9),
  951--958 (2014)

\bibitem{arbelle2018probabilistic}
Arbelle, A., Reyes, J., Chen, J.Y., Lahav, G., Riklin~Raviv, T.: A
  probabilistic approach to joint cell tracking and segmentation in
  high-throughput microscopy videos. Medical Image Analysis  \textbf{47},
  140--152 (2018)

\bibitem{arbelle2019microscopy}
Arbelle, A., Riklin~Raviv, T.: Microscopy cell segmentation via convolutional
  lstm networks. In: IEEE International Symposium on Biomedical Imaging (ISBI).
  pp. 1008--1012. IEEE (2019)

\bibitem{bensch2015cell}
Bensch, R., Ronneberger, O.: Cell segmentation and tracking in phase contrast
  images using graph cut with asymmetric boundary costs. In: IEEE International
  Symposium on Biomedical Imaging (ISBI). pp. 1220--1223. IEEE (2015)

\bibitem{bise2011reliable}
Bise, R., Yin, Z., Kanade, T.: Reliable cell tracking by global data
  association. In: IEEE International Symposium on Biomedical Imaging (ISBI).
  pp. 1004--1010. IEEE (2011)

\bibitem{blackman1986multiple}
Blackman, S.S.: Multiple-target tracking with radar applications. Dedham
  (1986)

\bibitem{braso2020learning}
Bras{\'o}, G., Leal-Taix{\'e}, L.: Learning a neural solver for multiple object
  tracking. In: IEEE Conference on Computer Vision and Pattern Recognition
  (CVPR). pp. 6247--6257 (2020)

\bibitem{chen2021celltrack}
Chen, Y., Song, Y., Zhang, C., Zhang, F., O’Donnell, L., Chrzanowski, W.,
  Cai, W.: Cell{T}rack {R}-{CNN}: {A} novel end-to-end deep neural network for
  cell segmentation and tracking in microscopy images. In: IEEE International
  Symposium on Biomedical Imaging (ISBI). pp. 779--782. IEEE (2021)

\bibitem{eom2018phase}
Eom, S., Sanami, S., Bise, R., Pascale, C., Yin, Z., Huh, S., Osuna-Highley,
  E., Junkers, S.N., Helfrich, C.J., Liang, P.Y., et~al.: Phase contrast
  time-lapse microscopy datasets with automated and manual cell tracking
  annotations. Scientific Data  \textbf{5}(1),  1--12 (2018)

\bibitem{Fey/Lenssen/2019}
Fey, M., Lenssen, J.E.: Fast graph representation learning with {PyTorch
  Geometric}. In: ICLR Workshop on Representation Learning on Graphs and
  Manifolds (2019)

\bibitem{gilmer2017neural}
Gilmer, J., Schoenholz, S.S., Riley, P.F., Vinyals, O., Dahl, G.E.: Neural
  message passing for quantum chemistry. In: International Conference on
  Machine Learning (ICML). pp. 1263--1272. PMLR (2017)

\bibitem{hayashida2019cell}
Hayashida, J., Bise, R.: Cell tracking with deep learning for cell detection
  and motion estimation in low-frame-rate. In: International Conference on
  Medical Image Computing and Computer-Assisted Intervention (MICCAI). pp.
  397--405. Springer (2019)

\bibitem{Hayashida_2020_CVPR}
Hayashida, J., Nishimura, K., Bise, R.: {MPM}: Joint representation of motion
  and position map for cell tracking. In: IEEE Conference on Computer Vision
  and Pattern Recognition (CVPR) (June 2020)

\bibitem{he2016deep}
He, K., Zhang, X., Ren, S., Sun, J.: Deep residual learning for image
  recognition. In: IEEE Conference on Computer Vision and Pattern Recognition
  (CVPR). pp. 770--778 (2016)

\bibitem{he2017cell}
He, T., Mao, H., Guo, J., Yi, Z.: Cell tracking using deep neural networks with
  multi-task learning. Image and Vision Computing  \textbf{60},  142--153
  (2017)

\bibitem{hernandez2018cell}
Hernandez, D.E., Chen, S.W., Hunter, E.E., Steager, E.B., Kumar, V.: Cell
  tracking with deep learning and the {V}iterbi algorithm. In: International
  Conference on Manipulation, Automation and Robotics at Small Scales (MARSS).
  pp.~1--6. IEEE (2018)

\bibitem{hoffer2015deep}
Hoffer, E., Ailon, N.: Deep metric learning using triplet network. In:
  International workshop on similarity-based pattern recognition. pp. 84--92.
  Springer (2015)

\bibitem{isensee2021nnu}
Isensee, F., Jaeger, P.F., Kohl, S.A., Petersen, J., Maier-Hein, K.H.:
  nn{U}-net: {A} self-configuring method for deep learning-based biomedical
  image segmentation. Nature methods  \textbf{18}(2),  203--211 (2021)

\bibitem{jug2016moral}
Jug, F., Levinkov, E., Blasse, C., Myers, E.W., Andres, B.: Moral lineage
  tracing. In: IEEE Conference on Computer Vision and Pattern Recognition
  (CVPR). pp. 5926--5935 (2016)

\bibitem{kausler2012discrete}
Kausler, B.X., Schiegg, M., Andres, B., Lindner, M., Koethe, U., Leitte, H.,
  Wittbrodt, J., Hufnagel, L., Hamprecht, F.A.: A discrete chain graph model
  for 3{D}+ t cell tracking with high misdetection robustness. In: European
  Conference on Computer Vision (ECCV). pp. 144--157. Springer (2012)

\bibitem{kingma2014adam}
Kingma, D.P., Ba, J.: Adam: A method for stochastic optimization. In:
  International Conference on Learning Representations (ICLR) (2015)

\bibitem{kipf2016semi}
Kipf, T.N., Welling, M.: Semi-supervised classification with graph
  convolutional networks. In: International Conference on Learning
  Representations (ICLR) (2017)

\bibitem{magnusson2015global}
Magnusson, K.E., Jald{\'e}n, J., Gilbert, P.M., Blau, H.M.: Global linking of
  cell tracks using the {V}iterbi algorithm. IEEE Transactions on Medical
  Imaging  \textbf{34}(4),  911--929 (2014)

\bibitem{mavska2013segmentation}
Ma{\v{s}}ka, M., Dan{\v{e}}k, O., Garasa, S., Rouzaut, A., Munoz-Barrutia, A.,
  Ortiz-de Solorzano, C.: Segmentation and shape tracking of whole fluorescent
  cells based on the {C}han--{V}ese model. IEEE Transactions on Medical Imaging
   \textbf{32}(6),  995--1006 (2013)

\bibitem{matula2015cell}
Matula, P., Ma{\v{s}}ka, M., Sorokin, D.V., Matula, P., Ortiz-de Sol{\'o}rzano,
  C., Kozubek, M.: Cell tracking accuracy measurement based on comparison of
  acyclic oriented graphs. PloS one  \textbf{10}(12),  e0144959 (2015)

\bibitem{10.1093/bioinformatics/btu080}
Maška, M., Ulman, V., Svoboda, D., Matula, P., Matula, P., Ederra, C.,
  Urbiola, A., España, T., Venkatesan, S., Balak, D.M., Karas, P., Bolcková,
  T., Štreitová, M., Carthel, C., Coraluppi, S., Harder, N., Rohr, K.,
  Magnusson, K.E.G., Jaldén, J., Blau, H.M., Dzyubachyk, O., Křížek, P.,
  Hagen, G.M., Pastor-Escuredo, D., Jimenez-Carretero, D., Ledesma-Carbayo,
  M.J., Muñoz-Barrutia, A., Meijering, E., Kozubek, M., Ortiz-de Solorzano,
  C.: {A benchmark for comparison of cell tracking algorithms}. Bioinformatics
  \textbf{30}(11),  1609--1617 (2014)

\bibitem{musgrave2020metric}
Musgrave, K., Belongie, S., Lim, S.N.: A metric learning reality check. In:
  European Conference on Computer Vision (ECCV). pp. 681--699. Springer (2020)

\bibitem{musgrave2020pytorch}
Musgrave, K., Belongie, S., Lim, S.N.: Py{T}orch metric learning (2020)

\bibitem{neumann2010phenotypic}
Neumann, B., Walter, T., H{\'e}rich{\'e}, J.K., Bulkescher, J., Erfle, H.,
  Conrad, C., Rogers, P., Poser, I., Held, M., Liebel, U., et~al.: Phenotypic
  profiling of the human genome by time-lapse microscopy reveals cell division
  genes. Nature  \textbf{464}(7289),  721--727 (2010)

\bibitem{nishimura2020weakly}
Nishimura, K., Hayashida, J., Wang, C., Bise, R.: Weakly-supervised cell
  tracking via backward-and-forward propagation. In: European Conference on
  Computer Vision (ECCV). pp. 104--121. Springer (2020)

\bibitem{padfield2011coupled}
Padfield, D., Rittscher, J., Roysam, B.: Coupled minimum-cost flow cell
  tracking for high-throughput quantitative analysis. Medical Image Analysis
  \textbf{15}(4),  650--668 (2011)

\bibitem{panteli2020siamese}
Panteli, A., Gupta, D.K., Bruijn, N., Gavves, E.: Siamese tracking of cell
  behaviour patterns. In: Medical Imaging with Deep Learning. pp. 570--587.
  PMLR (2020)

\bibitem{payer2019segmenting}
Payer, C., {\v{S}}tern, D., Feiner, M., Bischof, H., Urschler, M.: Segmenting
  and tracking cell instances with cosine embeddings and recurrent hourglass
  networks. Medical Image Analysis  \textbf{57},  106--119 (2019)

\bibitem{rempfler2017efficient}
Rempfler, M., Lange, J., Jug, F., Blasse, C., Myers, E.W., Menze, B.H., Andres,
  B.: Efficient algorithms for moral lineage tracing. In: IEEE International
  Conference on Computer Vision (ICCV). pp. 4695--4704 (2017)

\bibitem{rempfler2018tracing}
Rempfler, M., Stierle, V., Ditzel, K., Kumar, S., Paulitschke, P., Andres, B.,
  Menze, B.H.: Tracing cell lineages in videos of lens-free microscopy. Medical
  image analysis  \textbf{48},  147--161 (2018)

\bibitem{ronneberger2015u}
Ronneberger, O., Fischer, P., Brox, T.: U-net: {C}onvolutional networks for
  biomedical image segmentation. In: International Conference on Medical Image
  Computing and Computer-Assisted Intervention (MICCAI). pp. 234--241. Springer
  (2015)

\bibitem{rozemberczki2021pathfinder}
Rozemberczki, B., Englert, P., Kapoor, A., Blais, M., Perozzi, B.: Pathfinder
  discovery networks for neural message passing. In: Proceedings of the Web
  Conference. pp. 2547--2558 (2021)

\bibitem{scherr2020cell}
Scherr, T., L{\"o}ffler, K., B{\"o}hland, M., Mikut, R.: Cell segmentation and
  tracking using {CNN}-based distance predictions and a graph-based matching
  strategy. Plos One  \textbf{15}(12),  e0243219 (2020)

\bibitem{schiegg2015graphical}
Schiegg, M., Hanslovsky, P., Haubold, C., Koethe, U., Hufnagel, L., Hamprecht,
  F.A.: Graphical model for joint segmentation and tracking of multiple
  dividing cells. Bioinformatics  \textbf{31}(6),  948--956 (2015)

\bibitem{schiegg2013conservation}
Schiegg, M., Hanslovsky, P., Kausler, B.X., Hufnagel, L., Hamprecht, F.A.:
  Conservation tracking. In: IEEE International Conference on Computer Vision
  (ICCV). pp. 2928--2935 (2013)

\bibitem{sixta2020coupling}
Sixta, T., Cao, J., Seebach, J., Schnittler, H., Flach, B.: Coupling cell
  detection and tracking by temporal feedback. Machine Vision and Applications
  \textbf{31}(4),  1--18 (2020)

\bibitem{spilger2020recurrent}
Spilger, R., Imle, A., Lee, J.Y., Mueller, B., Fackler, O.T., Bartenschlager,
  R., Rohr, K.: A recurrent neural network for particle tracking in microscopy
  images using future information, track hypotheses, and multiple detections.
  IEEE Transactions on Image Processing  \textbf{29},  3681--3694 (2020)

\bibitem{sun2020circle}
Sun, Y., Cheng, C., Zhang, Y., Zhang, C., Zheng, L., Wang, Z., Wei, Y.: Circle
  loss: A unified perspective of pair similarity optimization. In: IEEE
  Conference on Computer Vision and Pattern Recognition (CVPR). pp. 6398--6407
  (2020)

\bibitem{ulman2017objective}
Ulman, V., Ma{\v{s}}ka, M., Magnusson, K.E., Ronneberger, O., Haubold, C.,
  Harder, N., Matula, P., Matula, P., Svoboda, D., Radojevic, M., et~al.: An
  objective comparison of cell-tracking algorithms. Nature methods
  \textbf{14}(12),  1141--1152 (2017)

\bibitem{velivckovic2017graph}
Veli{\v{c}}kovi{\'c}, P., Cucurull, G., Casanova, A., Romero, A., Li{\`o}, P.,
  Bengio, Y.: Graph attention networks. In: International Conference on
  Learning Representations (ICLR) (2018)

\bibitem{wang2020deep}
Wang, J., Su, X., Zhao, L., Zhang, J.: Deep reinforcement learning for data
  association in cell tracking. Frontiers in Bioengineering and Biotechnology
  \textbf{8}, ~298 (2020)

\bibitem{wang2019multi}
Wang, X., Han, X., Huang, W., Dong, D., Scott, M.R.: Multi-similarity loss with
  general pair weighting for deep metric learning. In: IEEE Conference on
  Computer Vision and Pattern Recognition (CVPR). pp. 5022--5030 (2019)

\bibitem{weng2020gnn3dmot}
Weng, X., Wang, Y., Man, Y., Kitani, K.M.: Gnn3dmot: Graph neural network for
  3d multi-object tracking with 2d-3d multi-feature learning. In: IEEE
  Conference on Computer Vision and Pattern Recognition (CVPR). pp. 6499--6508
  (2020)

\end{thebibliography}

\clearpage
\appendix

\section*{Appendices}
The supplementary material provides details about our implementation (Section~\ref{sec:imp_details}), as well as additional experiments used to evaluate our framework (Section~\ref{sec:exp}).
 

\section{Implementation Details}\label{sec:imp_details}
\subsection{Graph Neural Network}\label{subsec:graph_const}
We implemented the proposed graph neural network (GNN) model using the Pytorch Geometric library~\cite{Fey/Lenssen/2019}. 
We train our framework with graphs based on microscopy sub-sequences of $10$ frames while for the inference we use the entire sequence to construct the input graph. The prediction of all edges (a classification into `active' and `non-active' edges) is performed simultaneously. 

The spatio-temporal features are normalized by \textit{min-max scaling} for each graph, while the deep metric learning features are not pre-processed. 
To accommodate the high number of cell instances within a frame and to reduce the computational complexity, cell instances in consecutive frames are connected by edges only if their spatial Euclidean distance is smaller than a predefined threshold that is determined by the cells' \textit{neighborhood region}. 
The neighborhood region $\mathcal{N}_R$ is defined based on the size of the cells' bounding box $size_{BB}$ and the rate of the cells' movement $size_{move}$. 
Formally,
$\mathcal{N}_R = \alpha\cdot\max(\max_i(size_{BB_i}),\max_j(size_{move_j}))$. The maximization is applied to each axis separately. The hyper-parameter $\alpha$ is set to $2$ or $4$, depending on the sequence's density.
For the graph neural network, we set the number of layers $L=6$ to perform six message-passing steps, enabling information propagation between cell instances that are $6$ frames apart. 
We set the dimension $d_{\mathcal{V}}$ of the node feature matrix to $32$, where $d_{\mathcal{E}}=64$ for the edge feature matrix. The \textit{Adam} optimizer~\cite{kingma2014adam} is used with a learning rate of $1e-3$ and a weight decay of $1e-5$.

\subsection{Deep Metric Learning}
We use Pytorch metric learning library~\cite{musgrave2020pytorch} to train ResNet$18$~\cite{he2016deep} followed by multi-layer perceptron (MLP). The final embedding is $L2$ normalized and $d_{\mbox{\scriptsize DML}}=128$. 
The training is done using batches with a size of $32$. Batches are constructed by \textit{m-per-class} sampler, which first randomly samples $\kappa$ classes, and then randomly samples $m$ images for each of the $\kappa$ classes. 
Since the cell's appearance gradually changes during the sequence we perform the \textit{m-per-class} sampling~\cite{musgrave2020metric} using temporally adjacent frames. We set $\kappa = 8$ and $m = 4$.
The ResNet$18$ and MLP models are optimized using two separated \textit{Adam} optimizers~\cite{kingma2014adam} for each model with learning rates of $1e-5$ and $1e-4$, respectively. We also use weight decay of $1e-4$.
We use the cell segmentation maps or marker annotations to crop each frame into sub-images of all cell instances.
We constructed the datasets used for DML training by assigning to each cell instance the index of its biological cell.
In case the cell segmentation maps (rather than marker annotations) are available we exploit them to filter out the background via pixel-wise multiplication and extract features such as cell size and intensities.

\begin{figure*}[t]
    \centering 
    \subfigure[Control]{\label{fig:Control_viz}\includegraphics[width=58mm]{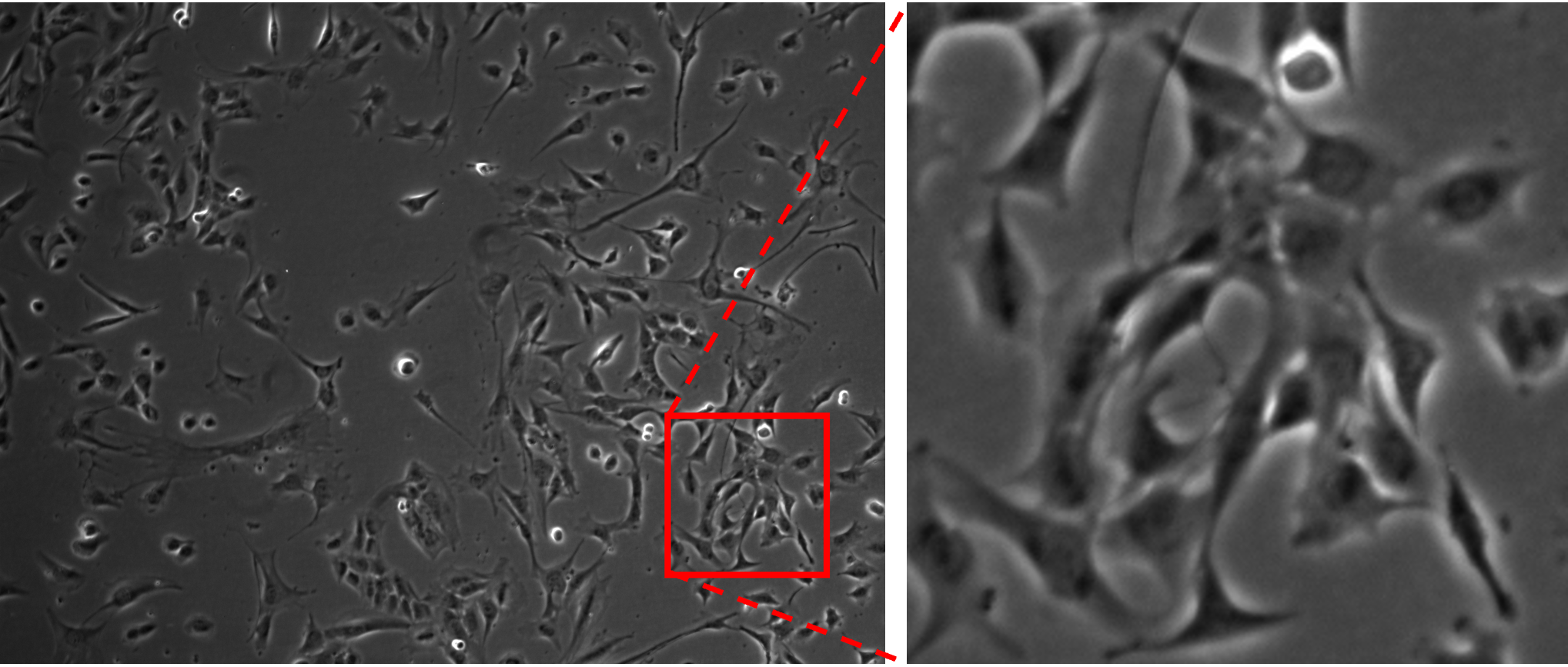}}\hskip 0.50cm
    \subfigure[FGF2]{\label{fig:FGF2_viz}\includegraphics[width=58mm]{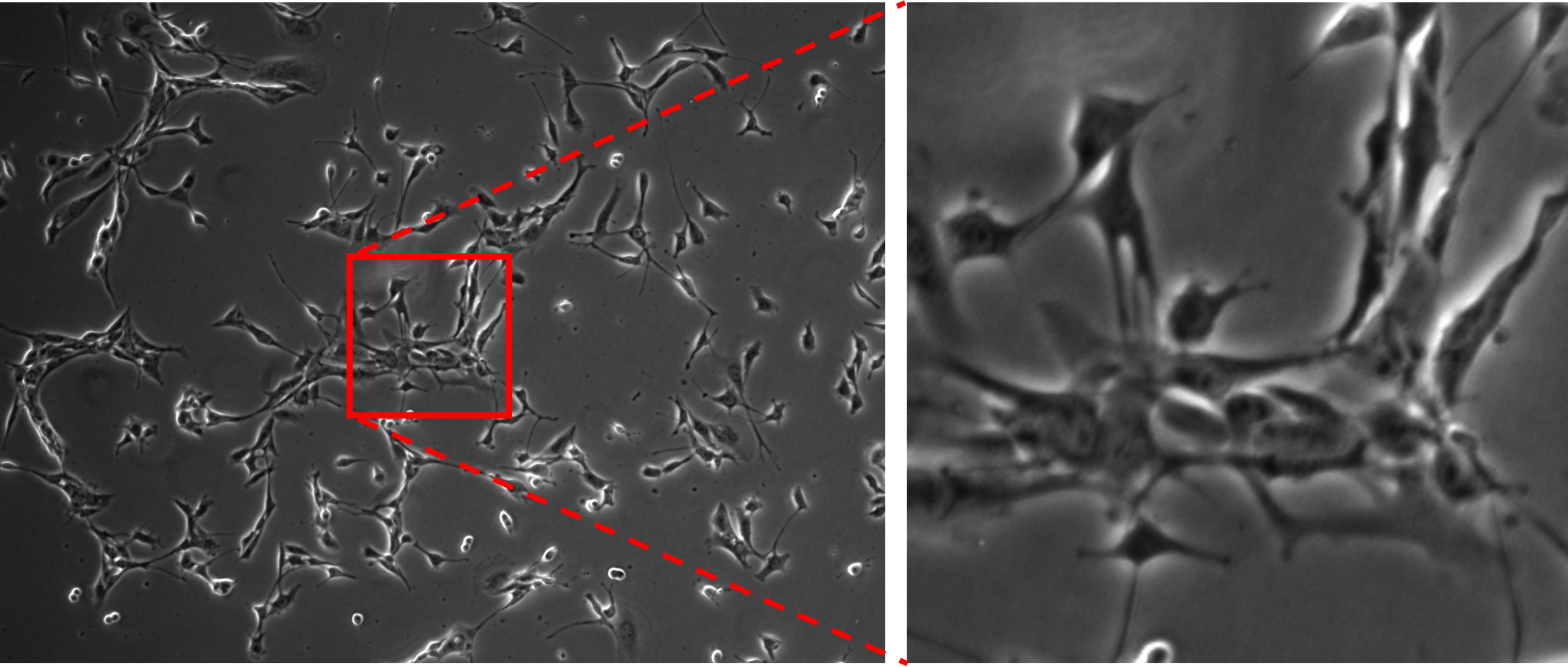}}\\
    \subfigure[BMP2]{\label{fig:BMP2_viz}\includegraphics[width=58mm]{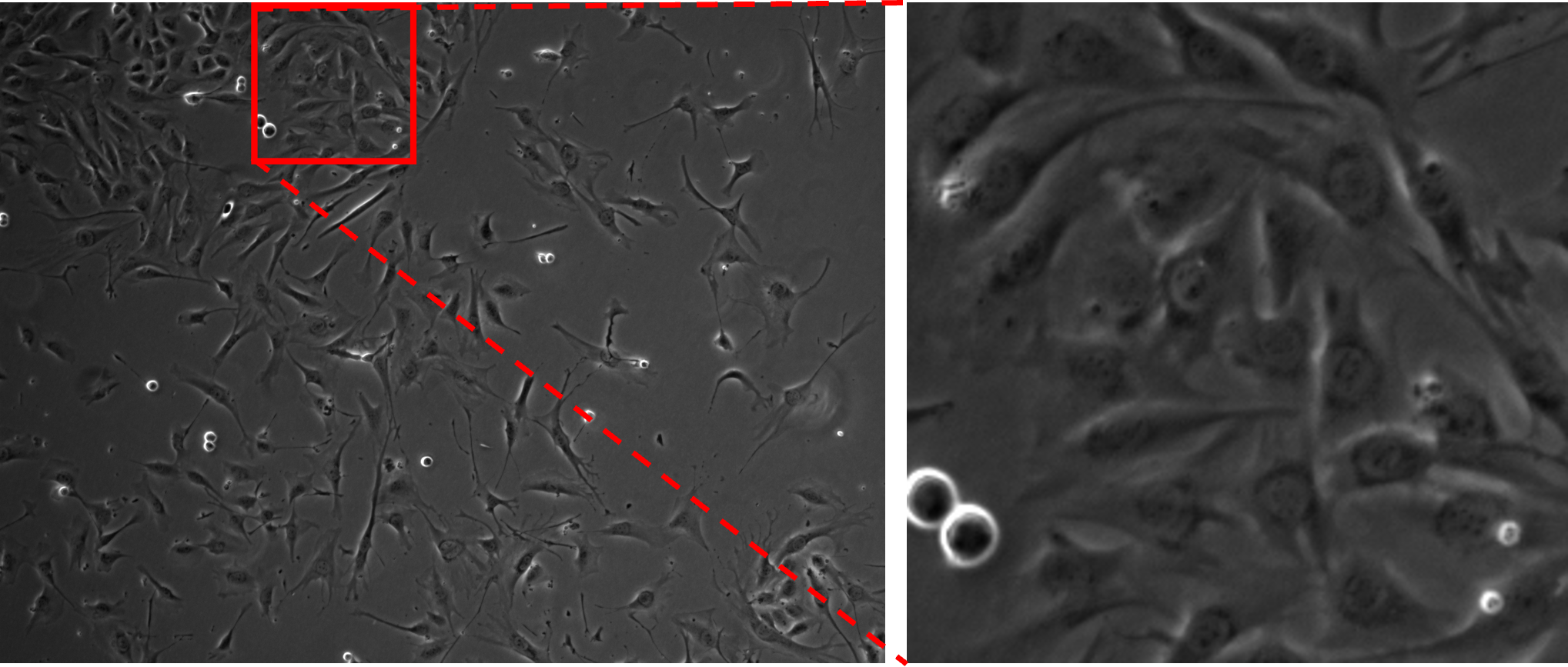}} \hskip 0.40cm
        \subfigure[FGF2 + BMP2]{\label{fig:FGF2_BMP2_viz}\includegraphics[width=58mm]{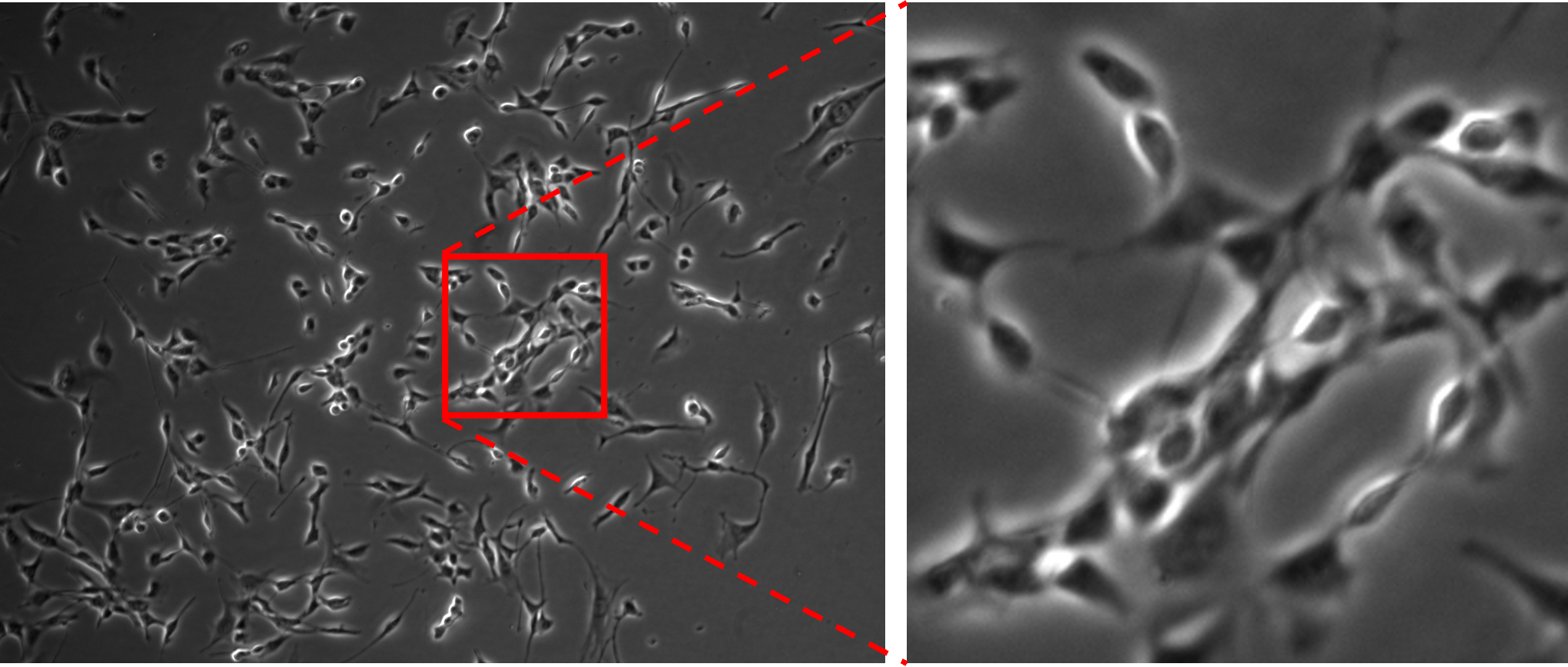}}
    \caption{\textbf{C2C12 Dataset Visualization.} Example frames of four C2C12 sequences with different growth factor conditions~\cite{eom2018phase}. Note the different appearance of the cells between the frames}
\label{fig:C2C12_visulavation}
\end{figure*}

\section{Experiments}\label{sec:exp}
We present additional experiments used to asses the proposed framework. In Section~\ref{sec:datasets_details}, we provide more details regarding the evaluated datasets. Then, 
in Section~\ref{subsec:qual_res}, we discuss the qualitative tracking results obtained for the C2C12 dataset~\cite{eom2018phase}. Comparison to graph-based solutions is summarized at Section~\ref{subsec:graph_compare}. Further ablation study experiments are presented in Section~\ref{sec:ablation_cont}. In Sections~\ref{subsec:diff} and~\ref{sec:complex}, we present an edge case and the run time of our framework, respectively. Last,
in Section \ref{subsec:dml_exp}, we
elaborate about the experiments used to evaluate our \textit{deep-metric-learning} based feature extractor.

\subsection{Datasets}\label{sec:datasets_details}
C2C12 dataset~\cite{eom2018phase} is acquired with four different growth factor conditions:fibroblast growth factor 2 (FGF2), bone morphogenetic protein 2 (BMP2), FGF2 + BMP2, and control (no growth factor). In Fig.~\ref{fig:C2C12_visulavation}, we present example frames for each condition. Note the different appearance of cells depending on the frames' growth factor.
The FGF2 cells are partially overlapped and become thinner along the sequence while the BMP2 cells are spread. In BMP2+FGF2, we can observe both phenomena.
In Table~\ref{tab:data_desc}, 
we summarize the properties of the CTC datasets~\cite{10.1093/bioinformatics/btu080,ulman2017objective} used to evaluate our method.

\begin{table}[t]

\setlength{\tabcolsep}{3pt}
  \centering
      \caption{Cell tracking challenge~\cite{10.1093/bioinformatics/btu080,ulman2017objective} datasets properties. The table provides details regarding the datasets dimension, cell type, acquisition method (phase-contrast or fluorescence microscopy), number of frames, and spatial resolution}
  \scalebox{1.0}{
    \begin{tabular}{lccccc}
    \midrule
    \midrule
    \textbf{Dataset} & \multicolumn{1}{l}{\textbf{Dim.}} & \multicolumn{1}{l}{\textbf{Cell Type}} & \multicolumn{1}{l}{\textbf{Acq.}} & \# \textbf{Frames}
     & \multicolumn{1}{l}{\textbf{Resolution}} \\
    \midrule
    \midrule
    PhC-C2DH-U373 & 2D    & U373  & Ph.-C. & 115   & $696 \times 520$ \\
    \midrule
    Fluo-N2DH-SIM+ & 2D    & HL60  & Fluo. & Varies & Varies \\
    \midrule
    Fluo-N3DH-SIM+ & 3D    & HL60  & Fluo. & Varies & Varies \\
    \midrule
    Fluo-C2DL-Huh7 & 2D    & HCC & Fluo. & 92    & $1024 \times 1024$ \\
    \midrule
    Fluo-N2DL-HeLa & 2D    & HeLa  & Fluo. & 30    & $1100 \times 700$ \\
    \bottomrule
    \end{tabular}}

  \label{tab:data_desc}%
\end{table}%

\subsection{Qualitative Tracking Results}\label{subsec:qual_res}
\begin{figure*}[h]
    \centering 
    \subfigure[Control]{\label{fig:Control}\includegraphics[width=58mm]{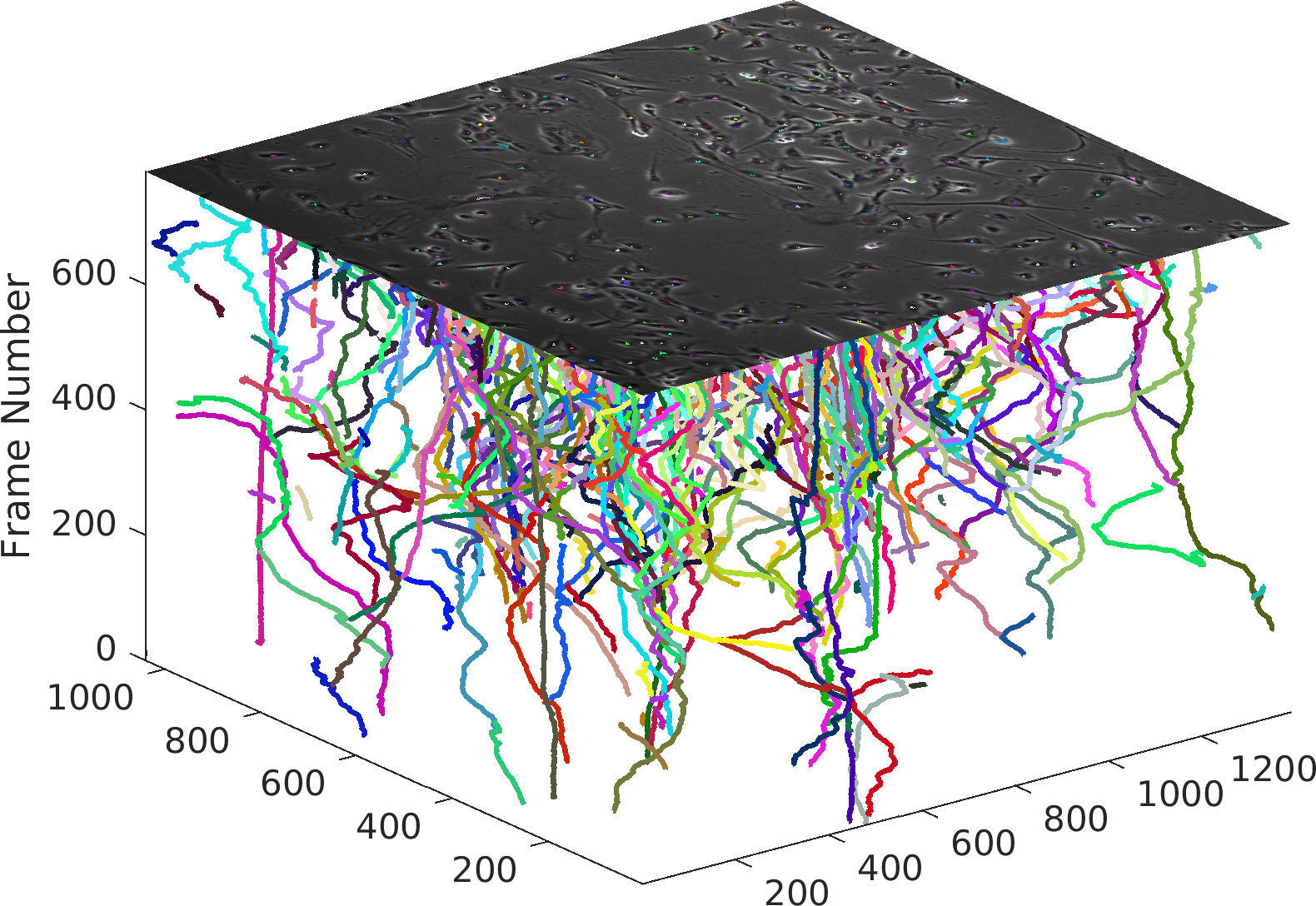}}\hskip 0.50cm
    \subfigure[FGF2]{\label{fig:FGF2}\includegraphics[width=58mm]{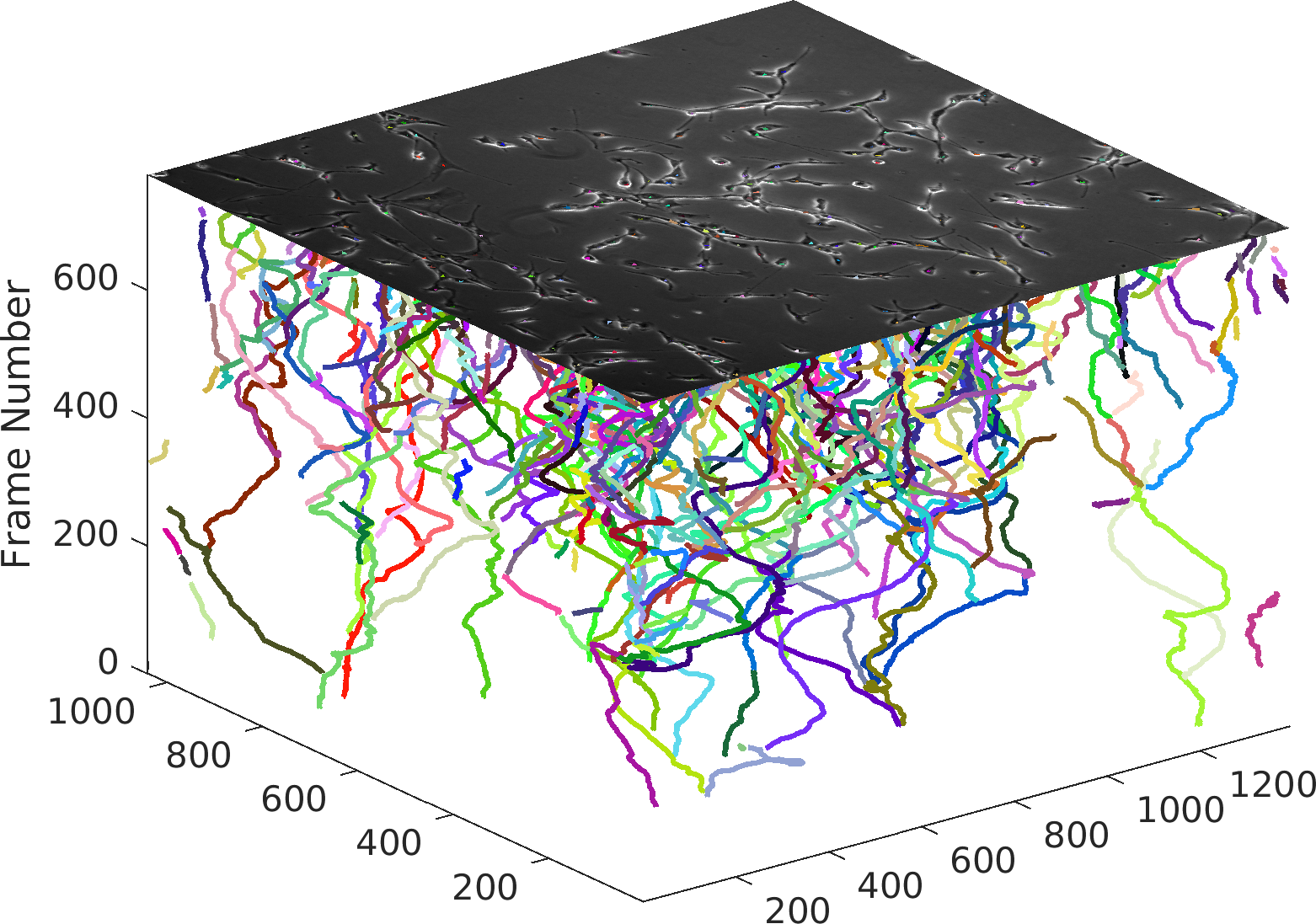}}\\
    \subfigure[BMP2]{\label{fig:BMP2}\includegraphics[width=58mm]{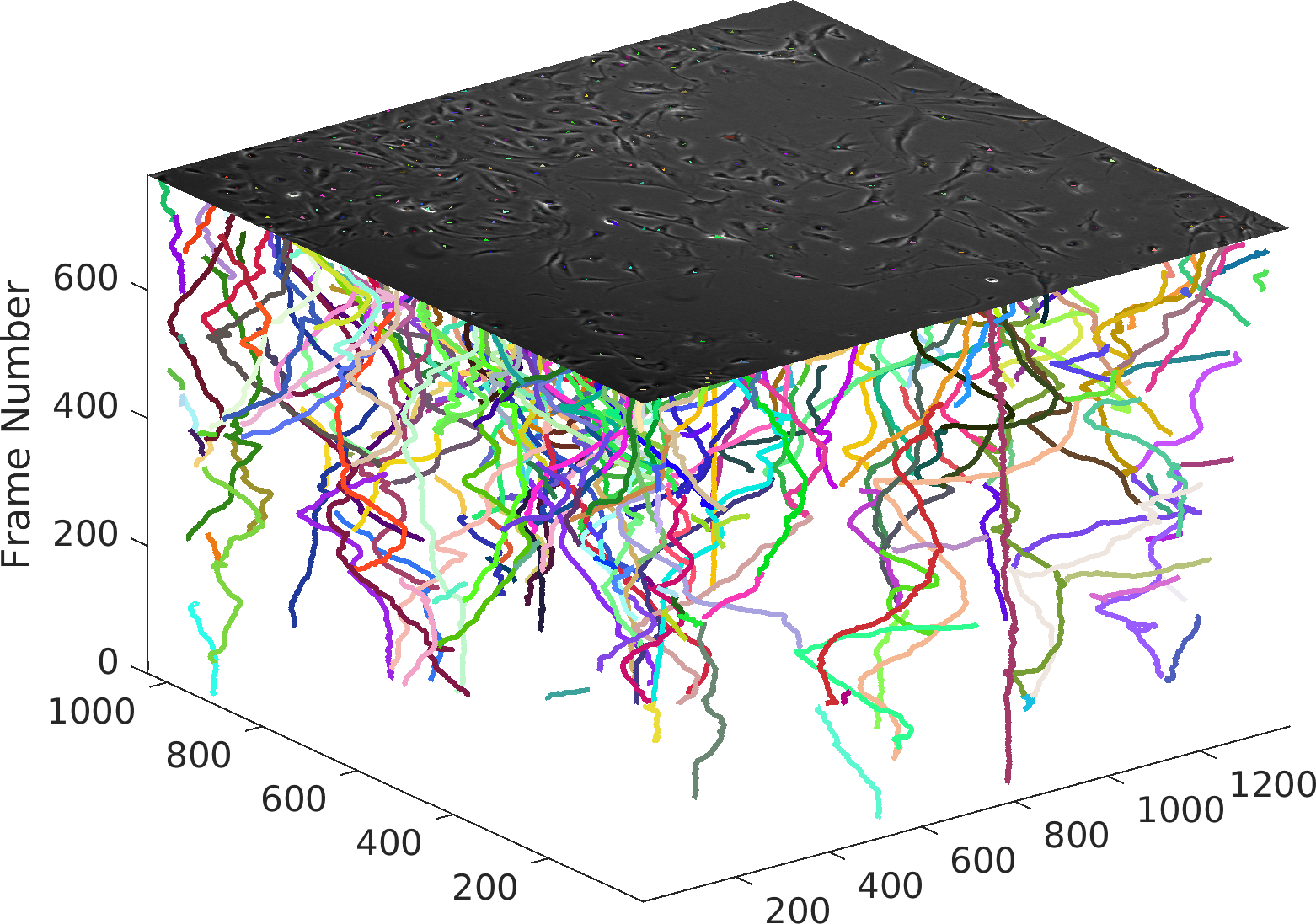}} \hskip 0.40cm
        \subfigure[FGF2 + BMP2]{\label{fig:FGF2BMP2}\includegraphics[width=58mm]{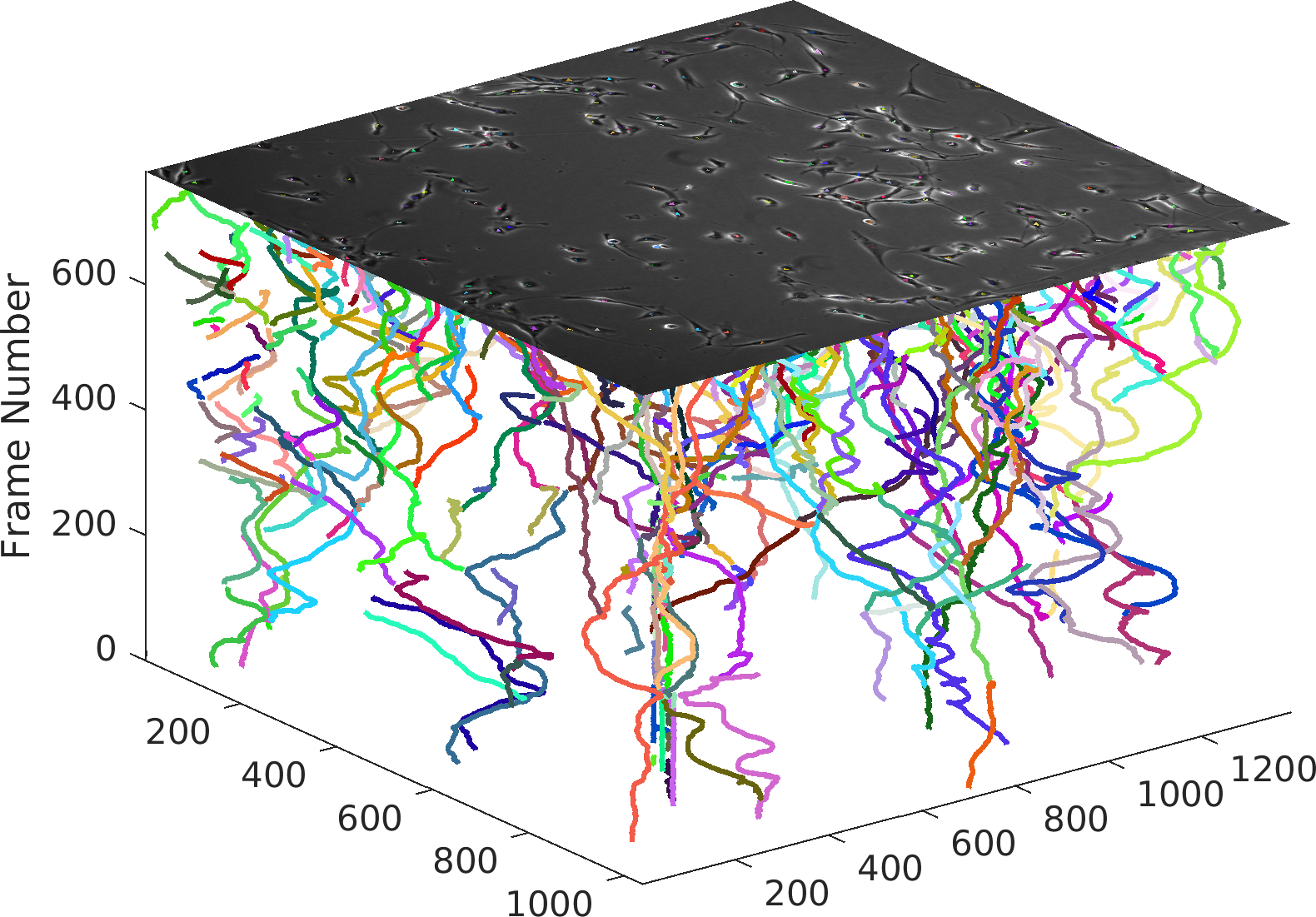}}
    \caption{\textbf{3D trajectories visualization.} Example trajectories of four different C2C12 sequences  dataset~\cite{eom2018phase}. The X-Y axes present the original frame resolution ($1392 \times 1040$ pixels), while Z axis presents the frame number. Note the clutter, density, overlap, and the random cell movements in each of the sequences}
\label{fig:tracks_visulavation}
\end{figure*}

Fig.~\ref{fig:tracks_visulavation} visually presents the trajectories of the C2C12 sequences. The figure illustrates the dense and cluttered cell environment and the prevalence of mitotic events and trajectory intersections. The video clips enclosed in this supplementary material further demonstrate the complexity of the extracted lineage trees and the strengths of the proposed method.

\subsection{Comparison to Graph-Based Methods}\label{subsec:graph_compare}
As mentioned in the main paper, we compare our method in Table~1 (main paper) with two classical graph-based methods, namely Asymmetric Graph Cut (AGC)~\cite{bensch2015cell}, Spatio-Temporal Global Data Association (ST-GDA)~\cite{bise2011reliable}.  Our method far surpasses AGC and ST-GDA, where the average results are improved by more than $15\%$ for the AA and the TE.
Moreover, a few algorithms which competed at the Cell Tracking Challenge (CTC) are based on a graph strategy. For example, Scherr et al.~\cite{scherr2020cell} adopt the coupled minimum-cost flow algorithm suggested in~\cite{padfield2011coupled}. Table~2 (main paper) presents our ranks for different CTC datasets with respect to all competing methods (three times top-1, one top-2, once top-4).

\subsection{Ablation Experiments - Cont.}\label{sec:ablation_cont}
We present more ablation experiments in Table~\ref{table:ablation} to demonstrate the contribution of the distance and the similarity features in the edge encoder to the overall cell tracking performance. 
As expected, the distance features are slightly better than the similarity, as more entries in the feature vector are represented by them. When using both features the results are more robust over all the datasets and improved. Last, by comparing first and last rows, the significant contribution of the distance similarity block is highlighted. The results demonstrate the exclusive and the joint contributions of the distance and the similarity features in the edge encoder to the overall cell tracking performance. 

\begin{table}[h!]
 \caption{\textbf{Edge encoder quantitative ablation study.} The contribution of the distance and the similarity features in the edge encoder to the overall cell tracking performance. The symbols \ding{51} and \ding{55} indicate included or excluded, respectively. AA and TE stand for association accuracy and target effectiveness scores }
\renewcommand{\arraystretch}{1.0}
\setlength{\tabcolsep}{8pt}
  \centering
  \scalebox{1.1}{
    \begin{tabular}{c|c|c|c}
    \hline
    \multicolumn{2}{c|}{Edge Encoder Components} & \multicolumn{2}{c}{Results} \\
    \hline
    Distance   & Similarity  & \multicolumn{1}{c|}{AA} & \multicolumn{1}{c}{TE} \\
    \hline
    \hline
    \ding{55}     & \ding{55}   & 0.840  & 0.598  \\
    \ding{55}    & \ding{51}    & 0.991  & 0.987  \\
    \ding{51}     & \ding{55}   & 0.993  & 0.988  \\
    \ding{51}  & \ding{51}  &  \textbf{0.994} & \textbf{0.989} \\
    \end{tabular}}
   \label{table:ablation}
\end{table}%

\subsection{Limitations and Edge Cases}\label{subsec:diff} 
\begin{figure*}[t!]
    \centering 
    \subfigure[Frame $t$]{\label{fig:Frame_t}\includegraphics[width=60mm]{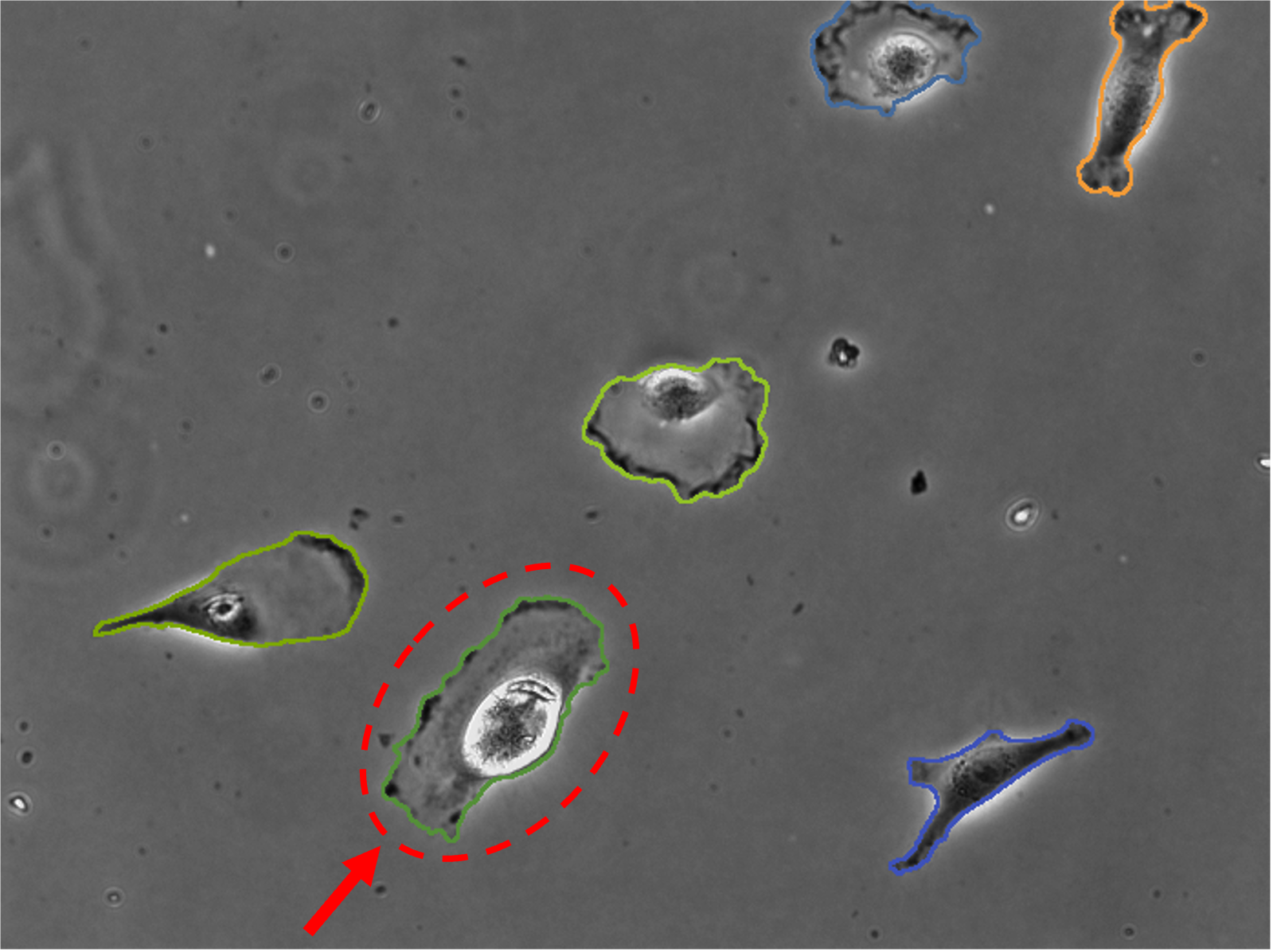}}
    \subfigure[Frame $t+1$]{\label{fig:Frame_t_plus}\includegraphics[width=60mm]{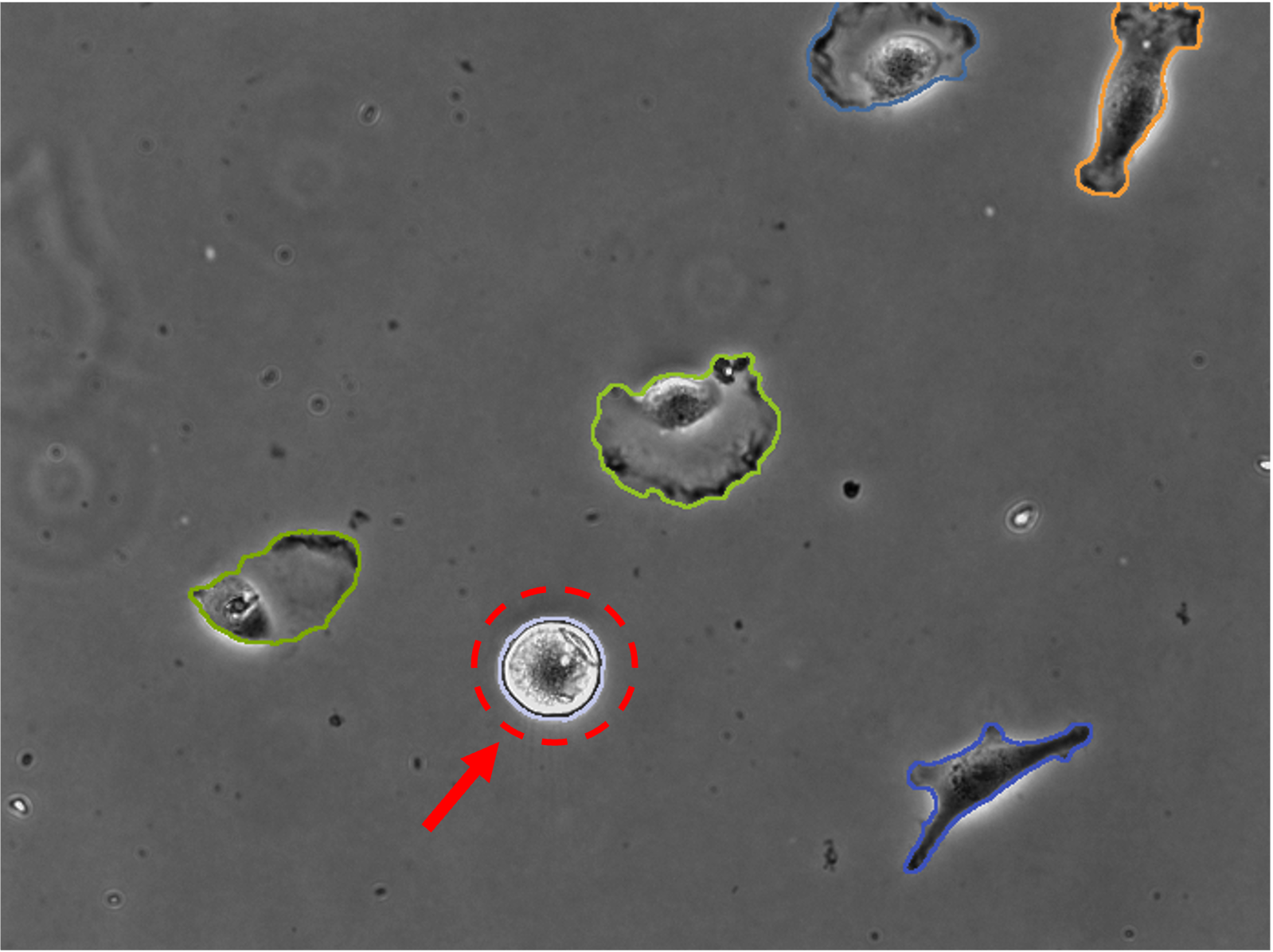}}
    \caption{\textbf{Edge case visualization.} Example of two consecutive frames from PhC-C2DH-U373 dataset. 
    The proposed framework may fail associating cell instances in consecutive frames when both cell dynamics and appearance change where the difference in either of them is significant and abrupt} 
\label{fig:Limitations}
\end{figure*}

The proposed tracker performs well even in the presence of abrupt changes in either cell appearance or dynamics (spatio-temporal changes). However, when both changes occur simultaneously the tracker may fail as illustrated in Fig.~\ref{fig:Limitations}. 

\subsection{Run time}\label{sec:complex}
We trained and tested our framework using NVIDIA Tesla V100 DGXS 32-GB GPU. Training and evaluation run times varied between a few minutes to a maximum of one hour depending on the sequence length (number of frames), cell density, and dimension (2D or 3D). Most of the time is spent on the construction of the graph. 
It is worth noting that cell tracking methods mainly run offline, after the acquisition of the entire time-lapse microscopy sequence. Therefore, the run time is not critical.

\subsection{Deep Metric Learning}\label{subsec:dml_exp}
In this section we present the experiments conducted to evaluate the DML component in our model.
\subsubsection{Evaluated Metrics.}\quad
For evaluation, we employ Adjusted Mutual Information (AMI) and Normalized Mutual Information (NMI), Precision at $1$ (P$@1$), R-Precision (RP), and Mean Average Precision at R (MAP$@R$)~\cite{musgrave2020metric} scores. 
AMI and NMI quantify the clustering performances and are based on $k$-mean algorithm, while P$@1$, RP, and MAP$@R$ measure the neighboring area and are based on $k$-nearest neighbors ($k$-nn) algorithm. Our primary evaluation metric is MAP$@R$ proposed recently in~\cite{musgrave2020metric}. This metric is stable and suited for the selection of the best performing model checkpoints.

\noindent\textbf{Ablation Study.}\quad
We conducted an ablation study to justify the use of \textit{multi-similarity loss} and miner~\cite{wang2019multi}. We compare the results with those obtained by the triplet loss~\cite{hoffer2015deep} with $L2$ distance and Cosine similarity in the embedded space. In both setting we use the \textit{triplet margin} miner. We also examine the recently proposed \textit{circle loss}~\cite{sun2020circle}. Finally, we compare the performances of the proposed sampling mechanism which selects within-class samples from temporally adjacent frames.

Table~\ref{tab:DML_ablation} presents the results obtained for each setup for the CTC Fluo-N2DL-HeLa dataset. We can observe that the \textit{multi-similarity loss} and the miner that was trained with the proposed sampler performs better than the other setups. Furthermore, the proposed modified sampling scheme outperforms the `traditional' \textit{$m$-per-class} sampler~\cite{musgrave2020metric} in a significant. This demonstrates its suitability to our task.

\begin{table}[t]
  \centering
  \caption{\textbf{Deep metric learning ablation study scores $(\%)$} }
  \setlength{\tabcolsep}{4pt}
   \scalebox{1.0}{
    \begin{tabular}{lcccc}
    \toprule
    \multicolumn{1}{l}{Method} & Sampler & P@1   & RP    & MAP@R \\
    \midrule
    Triplet+L2~\cite{hoffer2015deep} & Proposed & 80.6  & 33.2  & 27.4 \\
    Triplet+CS~\cite{hoffer2015deep} & Proposed & 82.3  & 35.1  & 29.7 \\
    Circle~\cite{sun2020circle} & Proposed & 83.1  & 36.8  & 31.7 \\
    MS~\cite{wang2019multi}   & $m$-per-class & 79.9  & 35.2  & 29.8 \\
    MS~\cite{wang2019multi}    & Proposed & \textbf{84.7}  & \textbf{37.8}  &\textbf{ 32.8} \\
    \bottomrule
    \end{tabular}}
  \label{tab:DML_ablation}
\end{table}%

\noindent\textbf{Results.}\quad
We report our results for all the datasets in Table~\ref{table:DML_RES}. The results differ due to the significant variability between the datasets. The better scores obtained for the 3D sequences Fluo-N3DH-SIM+ with respect to the 2D Fluo-N2DH-SIM+ sequences demonstrate the importance of the additional dimension for distinguishing between cell instances.

\begin{table}[h!]
  \centering
  \caption{\textbf{Accuracy measures $(\%)$} for the DML component. Performances on the evaluated metrics for each dataset under the same setting}
  \setlength{\tabcolsep}{4pt}
   \scalebox{1.0}{
    \begin{tabular}{lccccc}
    \toprule
    Dataset & \multicolumn{1}{l}{\textcolor[rgb]{ .129,  .129,  .129}{AMI}} & \multicolumn{1}{l}{\textcolor[rgb]{ .129,  .129,  .129}{NMI}} & P@1   & RP    & MAP@R \\
    \midrule
    Fluo-N3DH-SIM+ & 73.1  & 82.4  & 98.3  & 67.7  & 65.1 \\
    Fluo-N2DH-SIM+ & 55.6  & 73.2  & 84.8  & 42.9  & 38.1 \\
    PhC-C2DH-U373 & 45.9  & 46.8  & 81.6  & 51.6  & 39.3 \\
    Fluo-C2DL-Huh7 & 86.8  & 89.1  & 94.5  & 65.1  & 61.2 \\
    Fluo-N2DL-HeLa & 60.5  & 74.6  & 82.8  & 39.3  & 34.1 \\
    C2C12~\cite{eom2018phase} & 67.5  & 76.4  & 97.9  & 46.8  & 42.4 \\
    \bottomrule
    \end{tabular}}
     \label{table:DML_RES}
  \label{tab:addlabel}%
\end{table}%


\end{document}